\documentclass{article}

\usepackage{microtype}
\usepackage{graphicx}
\usepackage{subcaption}
\usepackage{booktabs} 
\usepackage{hyperref}
\usepackage[table]{xcolor}

\usepackage[preprint]{icml2026}
\makeatletter
\renewcommand{\ICML@preprint}{\textit{Preprint.}}
\makeatother
\hypersetup{pdfsubject={Research preprint},pdfkeywords={video quality assessment, attribution, multimodal evaluation}}
\usepackage{colortbl}
\colorlet{tablerowgray}{cyan!15}
\usepackage{multirow}
\usepackage{amsmath}
\usepackage{amssymb}
\usepackage{mathtools}
\usepackage{amsthm}
\usepackage{array}
\usepackage{makecell}
\usepackage{enumitem}
\usepackage{wrapfig}
\usepackage{tabularx}
\usepackage{array}
\newcolumntype{C}{>{\centering\arraybackslash}X}
\usepackage[capitalize,noabbrev]{cleveref}

\theoremstyle{plain}

\theoremstyle{definition}

\theoremstyle{remark}

\usepackage[disable,textsize=tiny]{todonotes}
\DeclareUnicodeCharacter{2192}{\ensuremath{\rightarrow}}

\icmltitlerunning{Q-Save: Scoring and Attribution for Generated Video Evaluation}

\begin{document}

\twocolumn[
  \icmltitle{Q-Save: Towards \underline{S}coring and \underline{A}ttribution for Generated \underline{V}ideo \underline{E}valuation}

  \icmlsetsymbol{equal}{*}

  \begin{icmlauthorlist}
    \icmlauthor{Xiele Wu}{equal,sch,comp}
    \icmlauthor{Mingtao Chen}{equal,comp}
    \icmlauthor{Zicheng Zhang}{sch}
    \icmlauthor{Yixian Liu}{comp}
    \icmlauthor{Yiming Liu}{comp}
    \icmlauthor{Jianxiang Lu}{comp}
    \icmlauthor{Jin Wang}{comp}
    \icmlauthor{Shushi Wang}{sch}
    \icmlauthor{Chunyi Li}{sch}
    \icmlauthor{Zhichao Hu}{comp}
    \icmlauthor{Lliu Yuhong}{comp}
    \icmlauthor{Xiongkuo Min}{sch}
    \icmlauthor{Guangtao Zhai}{sch}
    \icmlauthor{Xiaohong Liu}{sch,sch2}
  \end{icmlauthorlist}

  \icmlaffiliation{sch}{Shanghai Jiao Tong University}
  \icmlaffiliation{sch2}{Shanghai Innovation Institute}
  \icmlaffiliation{comp}{Hunyuan}

  \icmlcorrespondingauthor{Xiaohong Liu}{xiaohongliu@sjtu.edu.cn}

  \vskip 0.3in
]

\printAffiliationsAndNotice{\icmlEqualContribution}

\begin{abstract}
  Evaluating AI-generated video (AIGV) quality hinges on three crucial dimensions: visual quality, dynamic quality, and text-video alignment. While numerous evaluation datasets and algorithms have been proposed, existing approaches are constrained by two limitations: the absence of systematic definitions for evaluation dimensions, and the isolated treatment of the three dimensions in separate models. Therefore, we introduce Q-Save, a holistic benchmark dataset and unified evaluation model for AIGV quality assessment. The Q-Save dataset contains nearly 10,000 video samples, each annotated with Mean Opinion Scores (MOS) and fine-grained attribution explanations across the three core dimensions. Leveraging this attribution-annotated dataset, we train the proposed Q-Save model, which adopts the SlowFast framework to balance accuracy and efficiency, and employs a three-stage training strategy with Chain-of-Thought (COT) formatted data: Supervised Fine-Tuning (SFT), Grouped Relative Policy Optimization (GRPO), and a final SFT round for stability, to jointly perform quality scoring and attribution generation. Experimental results demonstrate that Q-Save achieves superior performance in AIGV quality prediction while providing interpretable justifications. Code and dataset will be released upon publication.
\end{abstract}

\section{Introduction}
  \label{sec:intro}

\begin{figure*}
\centering
\includegraphics[width=1.0\textwidth]{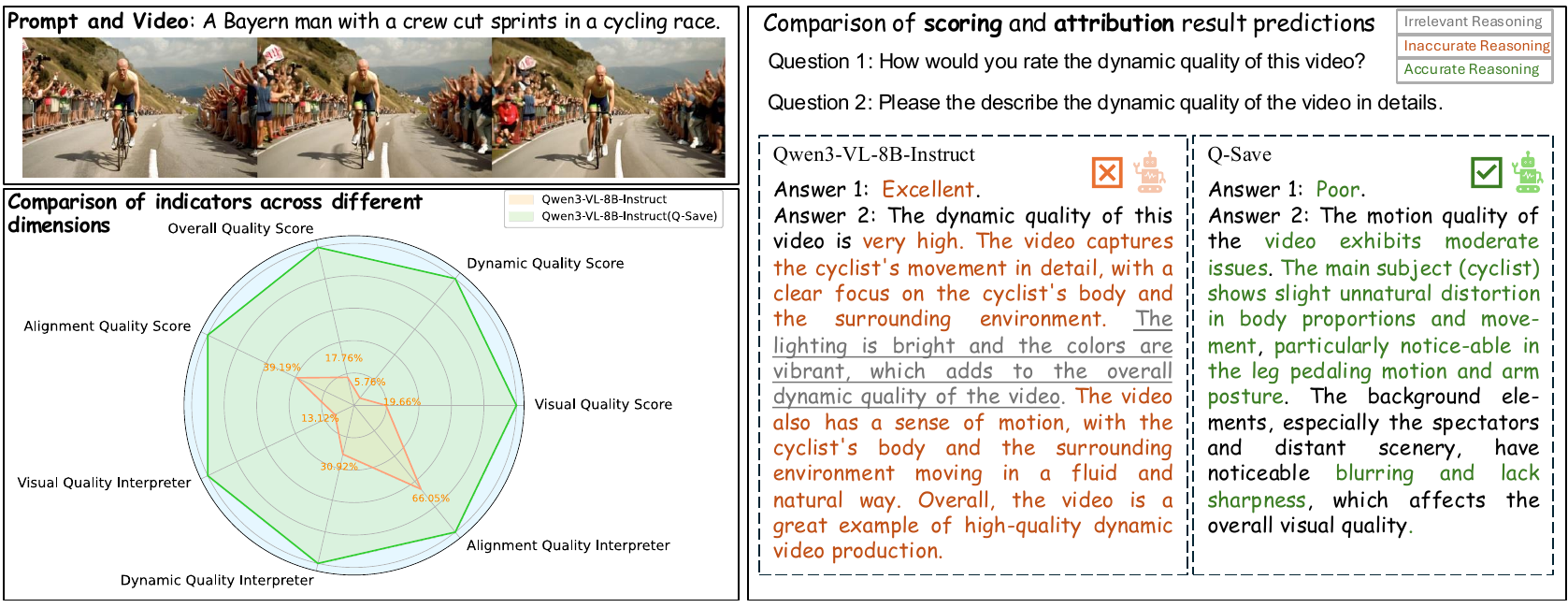}
\caption{Scoring and interpretation performance of Q-Save-tuned Qwen3-VL-8B-Instruct, compared to the baseline.} 
\label{spotlight}
\end{figure*}

  With the rapid advancement of generative AI, text-to-video (T2V) models are
producing and disseminating massive amounts of content across platforms for
  both industrial applications and end-user
  consumption~\cite{everypixel2024,techreport2024}. Yet, due to current
  technological limitations, most generated videos still exhibit varying degrees
  of imperfection (e.g., blur, artifacts, motion inconsistency, and prompt
  mismatch)~\cite{agiqa3k,vqascore,Q-Refine}. As a result, there is a growing
  demand for automated evaluation frameworks that can reliably assess the
  quality of generated videos and provide actionable feedback for model
  improvement~\cite{pickapic,imagereward,hpdv2,ku2023viescore,kim2023prometheus,zhang2023gpt,cho2023davidsonian,vqascore,q-instruct,qbench,wu2024qalign,zhang2024qboost}. Such frameworks can reduce the
  cost of manual review, improve evaluation efficiency, and serve as critical
  tools for guiding the optimization of generative models.

  To meet this need, many text-to-vision evaluation datasets and corresponding
  algorithms have been
  proposed~\cite{liu2024ntire,chen2024gaia,chen2023natural,he2024videoscore,huang2023vbench,agiqa3k,lmmpcqa,wang2023aigciqa2023,yuan2024pku,li2024aigiqa,richhumanfeedback,li2024genaibench,kou2024subjective,qbenchvideo,qbench+}. However,
  existing automated evaluation pipelines still face several practical
  limitations. First, prompts in many datasets are largely sampled or filtered
  from public sources (e.g., captions or existing benchmarks), which often leads
  to suboptimal prompt quality and an imbalanced distribution of content types. 
  Second, annotation quality control is frequently insufficient: rater onboarding and calibration are
  limited, and systematic auditing mechanisms are often missing, which increases
  noise and reduces the reliability of supervision. Third, most datasets provide
  only scalar scores without attributive explanations (i.e., \emph{why} a video
  is good or bad), which is increasingly inadequate for training high-quality,
  interpretable evaluators in complex multimodal
  settings~\cite{zhang2024survey}. Fourth, many VLM-based evaluation pipelines
  rely on overly sparse video preprocessing (e.g., sampling two frames per
  second), while human judgment of video quality is inherently grounded in
  richer temporal evidence beyond a few frames; this mismatch constrains model
  performance. Finally, common training recipes do not sufficiently exploit
  modern VLM capabilities: models are often trained with a single-stage
  supervised objective, leaving performance gains from stronger training
  strategies underexplored.

  In this work, we address these issues by building a high-quality, attribution-aware T2V evaluation dataset and a strong training pipeline for automated
  scoring. Based on extensive theoretical and empirical
  analyses~\cite{imagereward,hps,abench,zhang2024survey,liu2024ntire}, we define
  three fundamental dimensions for evaluating T2V generation:
  (1) \textbf{Visual Quality} measures frame-level perceptual quality (e.g.,
  fidelity, sharpness, artifacts, and aesthetics);
  (2) \textbf{Dynamic Quality} evaluates temporal dynamics (e.g., smoothness,
  coherence, and physical plausibility);
  (3) \textbf{Alignment with Text} assesses semantic consistency between the video
  content and the input prompt (i.e., generation accuracy). Based on these dimensions, we construct our dataset as follows:

\vspace{3pt}
  1) \textbf{Dataset construction.} We design prompts with a deliberate
  and balanced distribution over multiple coarse-grained categories and fine-grained subcategories, yielding a more reasonable coverage of content,
  actions, styles, and scene complexity. We generate videos using six state-of-the-art proprietary T2V models (at the time of collection) to ensure strong
  and diverse generation quality. (1) \textit{Annotation with strict quality
  control}. We recruit human raters and conduct multiple rounds of calibration
  to verify that each rater's understanding of
  the guidelines matches our requirements.
  (2) \textit{Attribution supervision}. For videos with noticeable defects, we
  further collect fine-grained attributions describing the causes of quality
  degradation along these three dimensions; our experiments show that such attribution data improves scoring
  precision and strengthens interpretability. To put Q-Save in context, Table~\ref{tab:dataset_benchmark_comparison} summarizes representative datasets and benchmarks for AIGV evaluation along several axes.
\begin{table}[t]
  \captionsetup{skip=3pt}
  \caption{Comparison of representative datasets/benchmarks for AIGV evaluation. \textbf{SBS} denotes side-by-side preference annotation and \textbf{MOS} denotes mean opinion score annotation.}
  \label{tab:dataset_benchmark_comparison}
  \centering
  \footnotesize
  \renewcommand{\arraystretch}{1.08}
  \setlength{\tabcolsep}{2pt}
  \begin{tabular*}{\columnwidth}{@{\extracolsep{\fill}}lcccccc@{}}
    \toprule
    \textbf{\makecell[l]{Dataset/\\Benchmark}} & \textbf{Year} & \textbf{\makecell{Closed\\Source}} & \textbf{\makecell{Annot.\\Type}} & \textbf{Scale} & \textbf{\#Dims} & \textbf{Attrib.} \\
    \midrule
    VideoScore2 & 2025 & No & MOS & 27k & 3 & Yes \\
    \makecell[l]{VideoGen-\\RewardBench} & 2025 & Yes & SBS & 108k & 4 & No \\
    T2VQA-DB & 2024 & No & MOS & 10k & 1 & No \\
    VideoPhy-2 & 2025 & No & MOS & 6.8k & 2 & No \\
    AIGVE-Bench & 2025 & No & MOS & 2.4k & 9 & No \\
    \midrule
    \textbf{Q-Save} & 2025 & Yes & MOS & 10k & 3 & Yes \\
    \bottomrule
  \end{tabular*}
\end{table}

\vspace{4pt}
  2) \textbf{Model and training.} To maximize evaluation capability, we build our
  evaluator on a strong video-capable VLM, \emph{Qwen3-VL-8B-Instruct}. Many VLM-based evaluation pipelines rely on overly sparse video preprocessing (e.g., sampling two frames per second), which constrains performance because human judgment is grounded in richer temporal evidence. We adopt a SlowFast-inspired strategy to capture both
  coarse temporal structure and fine-grained motion cues while balancing accuracy and efficiency. We also employ a three-
  stage training recipe to better exploit modern VLM capabilities beyond single-stage supervised training: (1) supervised fine-tuning (SFT) for a cold start, (2)
  reinforcement learning (RL) to further improve scoring performance and
  preference alignment, and (3) a final SFT stage to enhance stability and reduce variance.

  Our core contributions can be summarized as follows:

\begin{itemize}[itemsep=0pt, left=0pt, labelsep=5pt]
   \item We construct a \textbf{high-quality} T2V evaluation dataset with strict annotation
  quality control and attribution explanations that improve scoring precision
  and interpretability.
   \item We introduce a \textbf{SlowFast-inspired} video preprocessing strategy tailored for
  VLM-based evaluation to better leverage temporal evidence.
   \item We propose a \textbf{three-stage} training pipeline (SFT $\rightarrow $RL$ 
  \rightarrow$SFT) that more fully unlocks VLM capability and achieves state-of-the-art performance in AIGV evaluation.
\end{itemize}

\section{Related Work}
\label{gen_inst}

\subsection{Benchmarks for Text-to-Video Evaluation}

Early benchmarks for text-to-vision tasks primarily relied on multimodal datasets annotated with textual captions~\cite{coco_caption,ok-vqa,hu2023tifa}. However, as the importance of human feedback has gained increasing recognition, more recent benchmarks have shifted toward incorporating human-generated annotations~\cite{rlhf,zhang2024survey,huang2023t2i}. Two commonly adopted annotation strategies are Side-by-Side (SBS) comparisons and Mean Opinion Score (MOS) ratings~\cite{zerman2018relation,perez2019pairwise}. SBS involves selecting the preferred sample from a pair, offering higher annotation consistency and reduced cognitive load for annotators. In contrast, MOS assigns an absolute score to a single instance, offering greater flexibility and broader applicability across diverse evaluation contexts~\cite{leveque2021comparative,barratt2018note}. 

Evaluation dimensions in text-to-video tasks are generally grouped into three main categories~\cite{liu2024ntire,abench}: \textit{visual quality}, \textit{dynamic quality} and \textit{text-visual alignment}. Some benchmarks ~\cite{chen2024gaia,chen2023natural,he2024videoscore,huang2023vbench} define additional dimensions such as naturalness, aesthetics, and temporal coherence. Early image generation benchmarks~\cite{agiqa3k,wang2023aigciqa2023,yuan2024pku,li2024aigiqa} provided comprehensive evaluation protocols addressing both visual fidelity and semantic alignment. RichHF~\cite{richhumanfeedback} further advanced this by integrating subjective ratings, visual heatmaps, and misalignment tokens to offer more detailed insights. In the video domain, VideoFeedback~\cite{he2024videoscore} introduces a five-dimensional assessment framework covering both quality and alignment, whereas T2VQA-DB ~\cite{kou2024subjective} concentrates primarily on visual fidelity. Additionally, GenAI-Bench~\cite{li2024genaibench}  provides alignment evaluations for both image and video generation. Our work focuses on three core dimensions and pairs MOS ratings with attribution-style explanations, providing a unified framework for evaluating text-to-video generation.

\subsection{Reward Modeling for Vision}
Reward modeling has become a central paradigm for aligning generative models with human preferences in both image and video domains. Early methods such as DOVER~\cite{Dover} and ImageReward~\cite{imagereward} provide single scalar scores, which are effective but insufficient for capturing the multi-faceted nature of visual quality. More recent approaches such as VideoReward~\cite{VideoReward}, UnifiedReward~\cite{unifiedreward}, and Q-Insight~\cite{qinsight} introduce multi-dimensional scoring, yet are limited to numeric ratings without explanatory reasoning. Other efforts like LiFT~\cite{lift} provide short analytical comments, but remain broad and lack the depth necessary for systematic evaluation. Recent work such as VideoScore2~\cite{videoscore2} delivers multi-dimensional assessments together with long-form analytical reasoning, making its evaluations more interpretable and human-aligned. Our work complements these efforts by providing a unified benchmark and dataset to support fine-grained, reliable evaluation of generated videos.

\section{Data Construction}

\subsection{Basic Principles}

Our dataset is built through a multi-stage pipeline guided by three core principles: (1) \textbf{high-quality, human-written prompts} with multi-round refinement and quality checks; (2) \textbf{rigorous human quality control} via rater training, pilot rounds, and ongoing audits to ensure reliable annotations; and (3) \textbf{explanatory supervision for higher scoring precision}: in addition to MOS ratings, we collect natural language attribution explanations that acts as data augmentation to improve scoring precision (details in Appendix Sec.~\ref{sec:apdx_dataset_details}).

\begin{figure*}[h]
\centering
\includegraphics[width=1.0\textwidth]{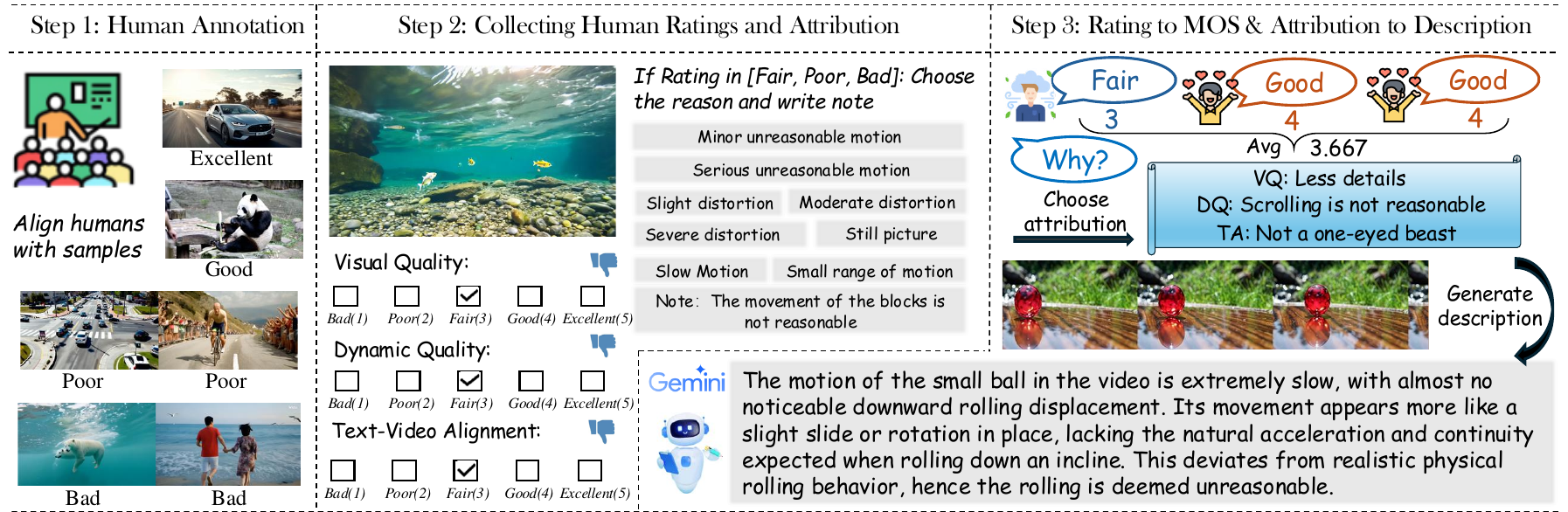}
\caption{Overview of the human annotation process. Step 1: raters are trained using text-defined rating levels (simulated via a rating-based syllabus for LMMs); Step 2: raters assign scores and select reasons for low ratings (Bad–Fair); Step 3: ratings are aggregated into MOS and reasons are converted into descriptions (via a probability-based inference method for LMMs).} 
\label{annotation}
\end{figure*}

\vspace{-3pt}
\subsection{Sources Collection}

\noindent\textbf{Prompt Collection.}\hspace{0.5em}Most existing video datasets derive prompts by sampling from pre-existing sources or filtering user-submitted content, which often yields ambiguous, underspecified, or low-quality prompts. In contrast, our prompts are \textbf{manually authored} to cover a comprehensive set of categories and to explicitly include dynamic actions and/or camera motions. Critically, we treat prompt quality as a first-class design goal: each prompt goes through multiple rounds of rewriting and human quality inspection to improve concreteness, compositional difficulty, and annotation friendliness. Detailed prompt taxonomy and statistics are provided in Appendix Sec.~\ref{subsec:apdx_dataset_construction}.

\noindent\textbf{Model Selection.}\hspace{0.5em}We conducted a systematic evaluation of both open-source and proprietary video generation models and selected six top-performing models based on preference rankings: Kling1.6~\cite{kling}, Kling2.0~\cite{kling}, Hunyuan~\cite{kong2025hunyuanvideosystematicframeworklarge}, Veo2, Dreamina~\cite{dreamina}, and Wanx~\cite{alibaba2024tongyiwanxiang}.

\subsection{Subjective Experiment}

\noindent\textbf{Annotation Process. }Given the critical importance of label quality for training evaluator models, we adopt a strict subjective annotation protocol. As illustrated in Figure~\ref{annotation}, all annotators complete a training program before formal annotation, followed by three rounds of pilot annotation to calibrate scoring criteria. During formal annotation, we conduct periodic manual inspections and remove annotators who fail to meet reliability and completeness requirements. We additionally collect attribution explanations describing why a video receives a particular score (see Appendix Sec.~\ref{subsec:apdx_annotation_protocol} for details). Beyond interpretability, this explanatory signal serves as supervision that improves scoring precision by grounding ratings in explicit evidence.

The dataset was divided into training and testing sets with an 80:20 split. Each training instance was annotated by at least three annotators, while each test instance received annotations from no fewer than twelve annotators to ensure high reliability in quality assessment. In total, our dataset comprises: 8,000 videos $\times$ 3 annotators $\times$ 4 dimensions $=96{,}000$ ratings for training, 2,000 videos $\times$ 12 annotators $\times$ 4 dimensions $=96{,}000$ ratings for testing, and nearly 40,000 natural language attribution explanations.

\vspace{2pt}
\subsection{Data format}

\vspace{1pt}
\subsubsection{Context Prompt}

\vspace{1pt}
Prior work on LMM-based evaluation often uses generic prompts lacking contextual grounding—for example, Q-Align~\cite{wu2024qalign} asks “Can you evaluate the quality of the image?”, while VQAScore~\cite{vqascore} uses binary yes/no questions. Such formulations are easy to implement but are often too vague for fine-grained and reliable scoring. To standardize the input format and improve scoring precision, we use a structured Chain-of-Thought (CoT) prompt that (1) explicitly specifies the evaluation dimension and criteria and (2) requests a brief attribution-style analysis before producing the final rating:

\vspace{-3pt}

\textit{Judge its \{dimension\} based on the following three criteria: \{dimension criteria\} . Considering the above video standards, you need to answer the following three questions: 1. Provide quality analysis of the video in this dimension. 2. How would you rate the visual quality of this video? Your rating can only be selected from the following five levels: [Bad, Poor, Fair, Good, Excellent]. Please integrate your answer with the following format: \textless think\textgreater Your analysis here\textless /think\textgreater Rating}

\vspace{-3pt}

This structured format encourages the model to ground scores in concrete evidence. During training, the resulting explanations are used as additional supervision, which acts as data augmentation and improves the accuracy and robustness of predicted ratings.

\subsubsection{Translating MOS into Ratings}
It is well established that discrete adjective ratings are easier for LLMs to interpret than numerical scores~\cite{wu2024qalign,zhang2024qboost}. Since MOS in Q-Save is labeled in absolute terms, we can easily map MOS to the corresponding rating:
\vspace{-5pt}
\[
R(s) = r_i \quad \text{if} \quad m + \frac{i - 1}{5}(M - m) < s \leq m + \frac{i}{5}(M - m),
\]

\vspace{-8pt}

\noindent where $\{r_i\}_{i=1}^5 = \{\text{Bad}, \text{Poor}, \text{Fair}, \text{Good}, \text{Excellent}\}$, $m = 1$ and $M = 5$ (the score range of Q-Save). $R(s)$ indicates the mapped rating of MOS value $s$.

\begin{figure*}[h]
\centering
\includegraphics[width=1.0\textwidth]{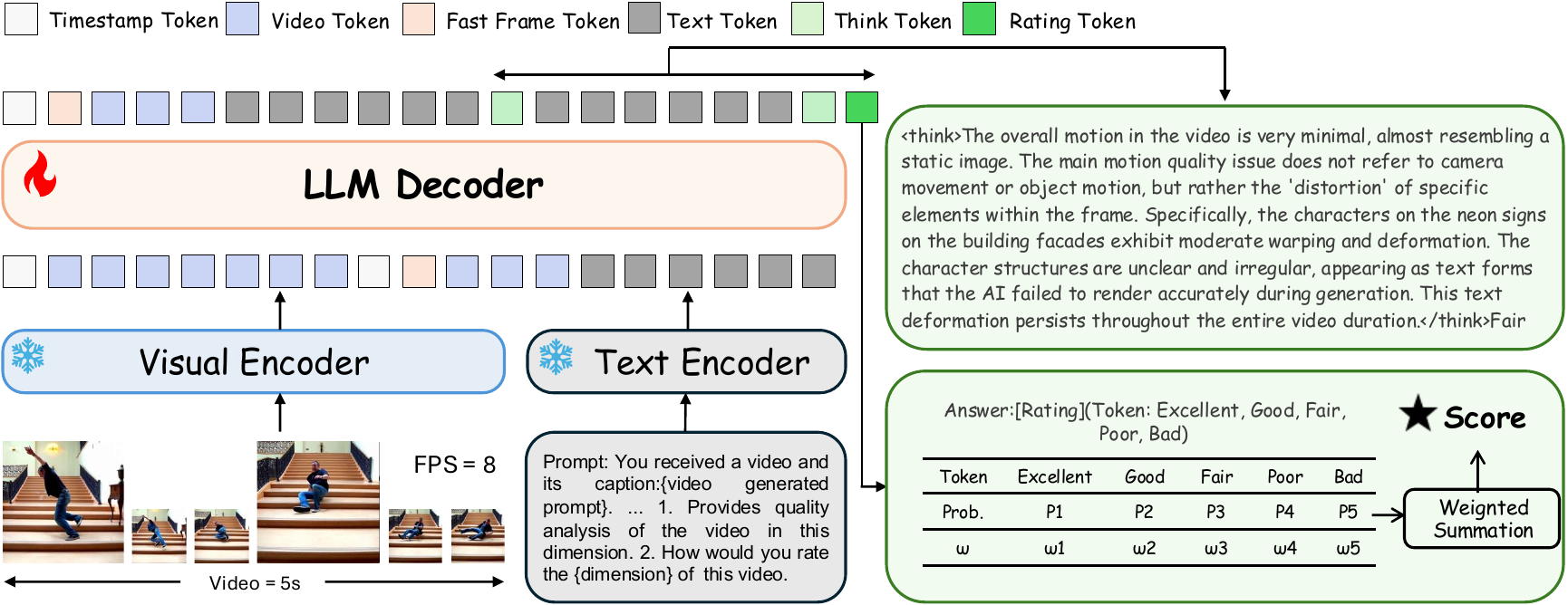}
\caption{Model architecture and video preprocessing. Our evaluator is based on a fine-tuned Qwen3-VL-8B-Instruct backbone and uses SlowFast-style preprocessing to extract spatiotemporal features (FPS = 8). } 
\label{model}
\end{figure*}

\section{Q-Save}

\subsection{Motivation and Design Choices}

\begin{figure}
\centering
\includegraphics[width=\columnwidth]{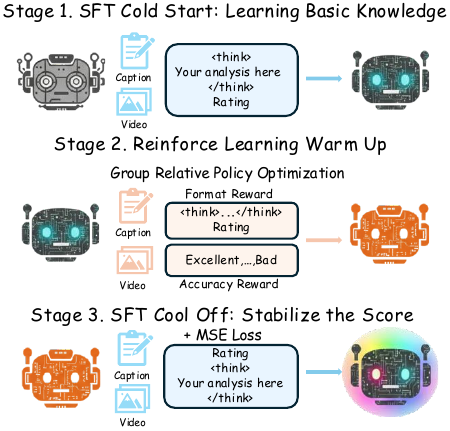}
\caption{Overview of the three-stage hybrid training pipeline. Blue denotes SFT data/models and orange denotes RL data/models (Cold Start → Warm-Up → Cool-Off).} 
\label{train_strategy}
\end{figure}

In this work, we develop a practical evaluator for AI-generated videos that is \textit{accurate}, \textit{robust}, and \textit{efficient} under the token-budget constraints of multimodal LLMs. We adopt Qwen3-VL-8B-Instruct~\cite{Qwen-VL} as the backbone because it offers a strong trade-off between video understanding capability, instruction-following, and computational efficiency. To address temporal redundancy in videos where adjacent frames are similar but key moments differ significantly, we employ a \textbf{SlowFast-style} preprocessing design that allocates more tokens to informative key frames (high-resolution ``slow'' frames) while still tracking temporal cues with lightweight ``fast'' frames, improving accuracy under a fixed token budget. For scoring, we predict discrete rating words (\textit{e.g.,} \{Excellent, Good, Fair, Poor, Bad\}), and applying a \textbf{softmax} over these candidate tokens yields a normalized distribution that captures uncertainty and supports stable continuous scoring via expectation. We combine \textbf{Cross-Entropy} and \textbf{MSE} losses to enforce both format correctness and numeric fidelity. For training, as shown in Figure~\ref{train_strategy}, we adopt a \textbf{three-stage} strategy: SFT provides a reliable cold start for instruction following; RL further aligns the model with evaluation objectives and mitigates shortcut behaviors; and a final SFT stage distills stable scoring behavior and reduces variance.

\vspace{1pt}
\subsection{Model Architecture}

We adopt Qwen3-VL-8B-Instruct~\cite{Qwen-VL} as the backbone LMM. As shown in Figure~\ref{model}, we modify \textit{the video preprocessing}, \textit{scoring method}, and \textit{training loss}.

\vspace{1pt}
\subsubsection{Video Processing}

\vspace{3pt}
For video encoding with varying FPS, resolutions, and durations, increasing any of these factors can sharply increase the token budget on the LLM side—making it challenging to balance performance and cost. To our knowledge, most existing MLLMs adopt a fixed number of frames and correspondingly reduce the resolution of each frame to meet token-budget constraints. Nevertheless, under uniform frame sampling, even though many adjacent frames may be highly similar, consecutive frames can still exhibit substantial differences. Considering the inherent characteristics of videos—where adjacent frames are mostly similar yet occasionally change significantly—we employ a SlowFast video encoding strategy~\cite{keye}: The \textbf{Slow Pathway} processes rapidly changing frames with fewer frames at higher resolution, while the \textbf{Fast Pathway} processes relatively static frames with more frames at lower resolution.

A patch-based similarity function is developed to identify slow/fast frames. The first frame is designated as a slow frame. For each subsequent frame, if its patch similarity with the most recent slow frame exceeds 90\%, it is labeled as a fast frame; otherwise, it is marked as a new slow frame. Special tokens with absolute timestamps are introduced to guide the model during training.

\subsubsection{Scoring Method}
The scoring calculation method is detailed below. We locate the rating token position and collect the logits corresponding to the five rating candidates \( \{Excellent, Good, Fair, Poor, Bad\} \). We then apply a softmax over these logits to obtain the normalized probabilities \(p_j\), where \( j \in \{1, 2, 3, 4, 5\} \). Finally, we compute a continuous predicted rating \( \hat{r} \) as the expectation under this distribution:

\vspace{-15pt}
\[
\hat{r} = \sum_{j=1}^{5} p_j \cdot w_j,
\]
\vspace{-15pt}

where \( w_j \) is the numerical weight assigned to each rating (\( w_j = \{1, 0.75, 0.5, 0.25, 0\} \) from Excellent to Bad).

\subsubsection{Loss function}

The loss consists of two complementary components: Cross-Entropy (CE) Loss and Mean Squared Error (MSE) Loss. CE trains the model to follow the desired response format and to generate the correct rating token, while MSE directly optimizes the continuous score to match MOS, improving numerical fidelity and calibration. The total loss is a weighted sum of the two.

The CE Loss helps the LMM learn the general question-answer format, defined as:
\vspace{-4pt}
\[
\mathcal{L}_{CE} = -\sum_{i=1}^{N} y_{i} \cdot \log(p_{i}),
\]
where \( y_{i} \) is the encoded vector of the true label for instance \( i \), and \( p_{i} \) is the predicted probability vector for the answer tokens. The MSE Loss refines the score prediction accuracy, given by:
\vspace{-4pt}
\[
\mathcal{L}_{MSE} = \left( \hat{r} - r_{MOS} \right)^2,
\]
where \( \hat{r} \) and \( r_{MOS} \) are the predicted score and the MOS label, respectively. The total loss is a weighted sum of the two:
\vspace{-4pt}
\[
\mathcal{L} = \alpha_{1} \cdot \mathcal{L}_{CE} + \beta_{1} \cdot \mathcal{L}_{MSE},
\]
with \( \alpha_{1} \) and \( \beta_{1} \) (default 1 and 1) controlling the contribution of each term.

\begin{table*}[t]
  \captionsetup{skip=3pt}
  \caption{Performance comparison on the Q-Save test set across visual quality, dynamic quality, and text-video alignment. Best in \textbf{bold}, second best \underline{underlined}.}
  \label{q_save_test_result}
  \centering
  \small
  \renewcommand{\arraystretch}{1.08}
  \setlength{\tabcolsep}{4pt}
  \begin{tabular}{lcccccccccccc}
    \toprule
      \multirow{3}{*}{\textbf{Model}} & \multicolumn{4}{c}{\textbf{Visual Quality}} & \multicolumn{4}{c}{\textbf{Dynamic Quality}} & \multicolumn{4}{c}{\textbf{Alignment}} \\
      \cmidrule(lr){2-5}\cmidrule(lr){6-9}\cmidrule(lr){10-13}
      & \multicolumn{2}{c}{\textit{Instance-level}} & \multicolumn{2}{c}{\textit{Model-level}} & \multicolumn{2}{c}{\textit{Instance-level}} & \multicolumn{2}{c}{\textit{Model-level}} & \multicolumn{2}{c}{\textit{Instance-level}} & \multicolumn{2}{c}{\textit{Model-level}} \\
      \cmidrule(lr){2-3}\cmidrule(lr){4-5}\cmidrule(lr){6-7}\cmidrule(lr){8-9}\cmidrule(lr){10-11}\cmidrule(lr){12-13}
     & SRCC & PLCC & SRCC & PLCC & SRCC & PLCC & SRCC  & PLCC & SRCC & PLCC & SRCC & PLCC \\
    \midrule
    FastVQA &0.087&0.062&-0.103&-0.091&0.125&0.073&-0.542&-0.376&0.294&0.271&0.513&0.432 \\
    SimpleVQA &0.152&0.119&-0.076&-0.048&0.193&0.138&0.089&0.021&0.316&0.295&0.547&0.479 \\
    Dover &0.238&0.214&0.169&0.157&0.286&0.263&0.227&0.201&0.352&0.334&0.569&0.513 \\
    Q-Align &0.275&0.253&0.198&0.184&0.312&0.291&0.254&0.232&0.378&0.359&0.583&0.536 \\
    VisionReward  &0.302&0.281&0.231&0.217&0.345&0.324&0.286&0.265&0.394&0.376&0.597&0.554 \\
    UnifiedReward &\underline{0.432}&\underline{0.415}&0.582&0.567&\underline{0.428}&\underline{0.409}&\underline{0.526}&\underline{0.513}&0.435&0.418&\underline{0.668}&\underline{0.654} \\
    VideoScore-v2 &0.397&0.381&\underline{0.613}&\underline{0.592}&0.395&0.378&0.498&0.482&\underline{0.447}&\underline{0.429}&0.643&0.627 \\
    \midrule
    \rowcolor{tablerowgray}
    \textbf{Q-Save} & \textbf{0.634} & \textbf{0.650} & \textbf{1.000} & \textbf{0.910} & \textbf{0.679} & \textbf{0.672} & \textbf{0.943} & \textbf{0.865} & \textbf{0.701} & \textbf{0.707} & \textbf{1.000} & \textbf{0.970} \\
    \bottomrule
  \end{tabular}
  
\end{table*}

\begin{table*}[t]
  \captionsetup{skip=3pt}
  \caption{Cross-dataset validation on out-of-domain benchmarks. \textbf{Left:} preference accuracy (\%) on pairwise benchmarks. \textbf{Right:} PLCC ($\times 100$) on MOS-style point-score benchmarks. \textbf{Bold} denotes the best model and \underline{underlined} denotes the second best. \textit{“/”} denotes the method trained on the benchmark's in-domain training set (excluded from comparison).}
  \label{cross_dataset_result}
  \centering
  \small
  \renewcommand{\arraystretch}{1.08}
  \setlength{\tabcolsep}{4pt}
  \begin{tabular*}{\textwidth}{@{\extracolsep{\fill}}lccc@{\hspace{10pt}}cccc@{}}
    \toprule
    \multirow{3}{*}{\textbf{Method}} & \multicolumn{3}{c}{\textbf{Preference Accuracy} (\%)} & \multicolumn{4}{c}{\textbf{MOS PLCC} ($\times 100$)}  \\
    \cmidrule(lr){2-4}\cmidrule(lr){5-8}
    & \multirow{2}{*}{\makecell{VideoGen\\RewardBench}} & \multirow{2}{*}{\makecell{GenAI\\Bench}} & \multirow{2}{*}{\makecell{T2VQA-DB\\Preference}} & \multirow{2}{*}{\makecell{VideoPhy2\\-test}} & \multicolumn{3}{c}{VideoScore-Bench-v2} \\
    \cmidrule(lr){6-8}
    &  &  &  &  &  Visual & Align & Phy \\
    \midrule
    VideoReward  & \textit{/} & 72.89 & 36.15 & 34.54 & 46.36 & 48.31 & 35.23     \\
    VisionReward & \underline{54.31} & 73.30 & 37.64 & 20.89 & 46.85 & 45.32 & \underline{38.25} \\
    Q-Align      & 42.05 & 64.25 & 43.24 & 4.11 & \textbf{54.71} & 34.01 & 37.78 \\
    AIGVE-MACS   & 37.09 & 53.21 & 36.91 & 10.00 & 27.30 & 6.90  & 13.03 \\
    VideoPhy2    & 30.75 & 51.50 & 24.12 & \underline{34.24} & 25.50  & 35.42 & 25.41 \\
    Dover        & 54.27 & 62.40 & 44.62 & 2.50 & 50.24 & 32.83 & 33.00 \\
    VideoScore-v2& 51.53 & \textbf{75.80} & \underline{50.60} & 33.58 & \textit{/} & \textit{/} & \textit{/} \\
    \midrule
    \textbf{Q-Save}& \textbf{56.88} & \underline{74.20} & \textbf{59.45} & \textbf{41.18} & \underline{49.25} & \textbf{58.70} & \textbf{38.57} \\
    \bottomrule
  \end{tabular*}
\end{table*}

\subsection{Training Strategy}

\noindent\textbf{SFT Cold Start. }We adopt a three-stage training strategy to balance format correctness, scoring accuracy, and stability. We first perform supervised fine-tuning (SFT) as the cold start to ensure basic format-following ability and task familiarity. 

\noindent\textbf{RL Warm Up. }We further train the SFT checkpoint using Group Relative Policy Optimization (GRPO), implemented in the open-source video reinforcement learning framework Verl~\cite{verl}. This stage strengthens alignment to evaluation objectives and improves robustness beyond supervised fitting. We use an additional smaller video dataset we constructed; details are provided in  Appendix Sec.~\ref{RL_dataset}. 

\textit{1) Accuracy Reward. }The accuracy reward is designed to directly guide the model to learn precise matching between its output and the true discrete labels of video reasoning tasks. The design principle of this reward signal is as follows: it is assigned a value of 1 if it matches the true discrete label in this dimension, and 0 if there is a mismatch. The formal definition is as follows: 

\vspace{-12pt}
\[
R_{\text{acc}} = 
\begin{cases}
1.0 & \text{if matching exactly}, \\
0 & \text{otherwise}.
\end{cases}
\]
\vspace{-12pt}

\textit{2) Format Reward. }To ensure the output includes both a rationale and a final rating, we assign \( R_{fmt} = 1 \) if the response contains the \textless think\textgreater \ tag with a rationale, and \( R_{fmt} = 0 \) otherwise.

\textit{3) Final Reward. }Following the setting for general video reasoning tasks in Video-R1~\cite{videor1}, the final reward is \( R = R_{acc} + \lambda R_{fmt}\), where \( \lambda = 0.5\).

\noindent\textbf{SFT Cool Off. }Finally, we distill stable scoring behavior by saving the correct rollouts, reformatting them into the structure of ``Rating\textless think\textgreater...reason...\textless/think\textgreater'', and running SFT again. This mitigates the instability sometimes introduced by RL and yields more consistent scores under our softmax-based scoring method.

\begin{table*}[t]
  \captionsetup{skip=3pt}
  \caption{Ablation study on training strategies. We report SRCC and PLCC at both instance-level and model-level across visual quality, dynamic quality, and text-video alignment. Best in \textbf{bold}, second best \underline{underlined}.}
  \label{tab:ablation}
  \centering
  \small
  \renewcommand{\arraystretch}{1.08}
  \setlength{\tabcolsep}{4pt}
  \begin{tabular*}{\textwidth}{@{\extracolsep{\fill}}lcccccccccccc@{}}
    \toprule
    \multirow{3}{*}{\textbf{Strategy}} & \multicolumn{4}{c}{\textbf{Visual Quality}} & \multicolumn{4}{c}{\textbf{Dynamic Quality}} & \multicolumn{4}{c}{\textbf{Alignment}} \\
    \cmidrule(lr){2-5}\cmidrule(lr){6-9}\cmidrule(lr){10-13}
    & \multicolumn{2}{c}{\textit{Instance-level}} & \multicolumn{2}{c}{\textit{Model-level}} & \multicolumn{2}{c}{\textit{Instance-level}} & \multicolumn{2}{c}{\textit{Model-level}} & \multicolumn{2}{c}{\textit{Instance-level}} & \multicolumn{2}{c}{\textit{Model-level}} \\
    \cmidrule(lr){2-3}\cmidrule(lr){4-5}\cmidrule(lr){6-7}\cmidrule(lr){8-9}\cmidrule(lr){10-11}\cmidrule(lr){12-13}
    & SRCC & PLCC & SRCC & PLCC & SRCC & PLCC & SRCC & PLCC & SRCC & PLCC & SRCC & PLCC \\
    \midrule
    Stage 1 w/o SlowFast & 0.50 & 0.51 & 0.58 & 0.57 & 0.58 & 0.57 & 0.54 & 0.58 & 0.63 & 0.61 & 0.79 & 0.75 \\
    Stage 1              & 0.57 & 0.54 & 0.61 & 0.60 & 0.61 & 0.60 & 0.52 & 0.51 & 0.66 & 0.65 & 0.84 & 0.81 \\
    Stage 1+2            & \underline{0.61} & \underline{0.60} & \underline{0.71} & \underline{0.63} & \underline{0.64} & \underline{0.63} & \underline{0.74} & \underline{0.65} & \underline{0.67} & \underline{0.66} & \underline{0.87} & \underline{0.86} \\
    \midrule
    Stage 1+2+3          & \textbf{0.63} & \textbf{0.65} & \textbf{1.00} & \textbf{0.91} & \textbf{0.68} & \textbf{0.67} & \textbf{0.94} & \textbf{0.86} & \textbf{0.70} & \textbf{0.70} & \textbf{1.00} & \textbf{0.97} \\
    \bottomrule
  \end{tabular*}
\end{table*}

\begin{table}[t]
  \captionsetup{skip=3pt}
  \caption{Human evaluation scores after reinforcement learning fine-tuning of video generation models in the game-CG scenario. Best in \textbf{bold}, second best \underline{underlined}. HPSv3 is an open-source reward model.}
  \label{tab:rl_video_gen}
  \centering
  \small
  \renewcommand{\arraystretch}{1.08}
  \setlength{\tabcolsep}{6pt}
  \begin{tabular*}{\linewidth}{@{\extracolsep{\fill}}lcc@{}}
    \toprule
    \textbf{Training Configuration} & \textbf{Wan2.2} & \textbf{Hunyuan1.5-video} \\
    \midrule
    w/o RL & 66.76 & 55.86 \\
    RL with HPSv3 & \underline{68.10} & \underline{56.23} \\
    RL with Q-Save-v1.5 (ours) & \textbf{69.95} & \textbf{56.68} \\
    \bottomrule
  \end{tabular*}
\end{table}

\section{Experiments}

\subsection{Experiment Setup}
\label{subsec:exp_setup}

\noindent \textbf{Training \& Evaluation. }We use Qwen3-VL-8B-Instruct ~\cite{Qwen-VL} as the backbone vision-language model (VLM) throughout training. All experiments were conducted on NVIDIA H20 GPUs.

\noindent \textbf{Implementation Details. }Key hyperparameters and inference settings are provided in Appendix Sec.~\ref{sec:apdx_impl_details}.

For evaluation metrics, we report correlation metrics (SRCC/PLCC under both \textbf{instance-level} and \textbf{model-level} protocols) on MOS-style benchmarks and accuracy on preference benchmarks. Detailed definitions are provided in Appendix Sec.~\ref{sec:apdx_eval_metrics}.

\noindent \textbf{Competitors. }We compare against representative video quality assessment and visual reward models, including FastVQA~\cite{liu2022fastvqa}, SimpleVQA~\cite{simpleVQA}, Q-Align~\cite{wu2024qalign}, DOVER~\cite{Dover}, VideoScore-v2~\cite{videoscore2}, UnifiedReward~\cite{unifiedreward}, VideoReward~\cite{VideoReward}, and VisionReward~\cite{visionreward}. Brief descriptions of these reward-model baselines are provided in Appendix Sec.~\ref{sec:apdx_reward_model_baselines}.

All models are evaluated using their publicly available implementations, recommended hyperparameters, and the corresponding splits of the Q-Save dataset unless otherwise specified.

\subsection{Discussion \& General Findings}
\label{subsec:discussion_findings}

Table~\ref{q_save_test_result} reports in-domain performance on our test split across three dimensions. Rather than only reflecting model scale, the consistent gains are mainly driven by how our \textbf{dataset design}, \textbf{video evidence allocation}, and \textbf{training strategy} match common failure modes of AIGC videos.

\noindent\textbf{Observation 1: AIGC-specific quality signals.} Classical no-reference VQA models are largely optimized for distortion patterns in natural/UGC videos, while AIGC videos often fail at a different level: object integrity, compositional plausibility, action completeness, and prompt-conditioned semantics. Because Q-Save couples each video with a carefully written prompt and evaluates on dimension-specific criteria, our model can use the prompt as contextual supervision to disambiguate alignment. 

\vspace{3pt}
\textbf{Observation 2: Model-Level Evaluation Strongly Improves Performance} In Q-Save, prompts are shared across multiple generators, so a strong evaluator must be \emph{calibrated} across diverse content and not drift with prompt difficulty; softmax-based score expectation reduces discretization noise and preserves uncertainty, while the multi-stage training reduces variance introduced by direct regression and improves consistency across prompts—particularly important for ranking generators (model-level), where small systematic biases accumulate across many samples. 

\vspace{3pt}
\textbf{Observation 3: Temporal evidence allocation for dynamic quality.} Dynamic quality errors are often temporally sparse (e.g., brief discontinuities or implausible motion), and SlowFast-style preprocessing focuses capacity on key-changing moments while still tracking motion context, improving sensitivity to these sparse artifacts under a fixed token budget; this design aligns with our data collection, where prompts intentionally include dynamic actions/camera motions, making temporal faithfulness a first-class signal rather than a byproduct of frame aesthetics. 

\textbf{Observation 4: Explanation-augmented supervision as data augmentation.} Our attribution explanations act as additional supervision that anchors MOS ratings to evidence. This reduces ambiguity in borderline cases and discourages shortcut learning, resulting in more transferable quality signals across visual, dynamic, and alignment dimensions.

\noindent \textbf{Cross-dataset Validation.}
To test transferability across evaluation paradigms, Table~\ref{cross_dataset_result} reports \textbf{preference accuracy} (left) and \textbf{MOS PLCC} (right). Benchmark definitions, dimension mappings, and rescaling rules are provided in Appendix Sec.~\ref{subsec:apdx_eval_benchmarks}. Strong performance in both blocks indicates that our evaluator learns a calibrated continuous scoring function: it preserves linear trends for MOS correlation and can be reliably thresholded to make preference decisions. In particular, good transfer on motion/physics-focused tests suggests that our video preprocessing and training strategy improve sensitivity to temporally sparse but decisive artifacts, which are often missed by frame-only scoring. Overall, the results support that explanation-augmented supervision in Q-Save acts as effective data augmentation for learning transferable reward signals. Full evaluation results are provided in Appendix Sec.~\ref{sec:apdx_full_eval_results}.

\noindent \textbf{Effect of Q-Save as a Reward Model on Video Generators.}
Beyond evaluation, Q-Save can be directly repurposed as a reward model to improve text-to-video generators. We conduct RL on two generators, Wan2.2~\cite{alibaba2024tongyiwanxiang} and Hunyuanvideo-1.5~\cite{wu2025hunyuanvideo15technicalreport}, and compare against both the non-RL baselines and RL using the open-source reward model HPSv3~\cite{ma2025hpsv3widespectrumhumanpreference}. As shown in Table~\ref{tab:rl_video_gen}, Q-Save-based RL yields clear gains in human evaluation, while HPSv3-based RL brings smaller improvements. Overall, Q-Save-based RL outperforms HPSv3-based RL on both generators, suggesting that Q-Save provides a more transferable optimization signal under the same RL setup. More details are provided in Appendix Sec.~\ref{subsec:apdx_rl_human_eval}

\subsection{Ablation Study}

To dissect the efficacy of our multi-stage training strategy—SFT cold-start (with/without SlowFast preprocessing), GRPO-based RL warm-up, and SFT cool-off—we conduct ablations in Table~\ref{tab:ablation}. Based on these ablation results, we draw the following conclusions: 
1) \noindent\textbf{SlowFast helps early evidence quality.} Stage 1 with SlowFast improves instance-level correlations across all three dimensions, consistent with the motivation that allocating more capacity to informative ``slow'' frames improves both spatial defect recognition and motion understanding under a fixed token budget.
2) \noindent\textbf{RL warm-up improves global calibration.} The largest gains appear after Stage 1+2, especially on model-level metrics. This suggests GRPO improves consistency across generators (model-level ranking), which is essential for evaluator/reward use cases such as model selection and best-of-$n$ sampling.
3) \noindent\textbf{SFT cool-off consolidates stability.} The final SFT stage further improves PLCC and yields the most stable model-level rankings, consistent with ``cool-off'' acting as a distillation step that reduces variance and format drift introduced by RL while preserving reward-aligned decision boundaries. Overall, the ablation supports that preprocessing (better evidence), RL (better alignment), and final SFT (better stability) are complementary components. 

\section{Conclusion}

In this paper, we present Q-Save, a large-scale dataset and evaluator for AI-generated video assessment. Building on Q-Save, we develop a unified evaluator based on Qwen3-VL-8B-Instruct with SlowFast-style video preprocessing, softmax-based score expectation, and a CE+MSE objective. A three-stage training pipeline yields stable and calibrated predictions. Experiments show strong in-domain performance and robust transfer across both preference-style and MOS-style benchmarks. Moreover, our evaluator can be directly repurposed as a reward model for optimizing video generators. Limitation are provided in Appendix Sec.~\ref{sec:limitations_and_future_work}

\newpage

\section{Impact Statement}

Q-Save provides reliable and diagnostic AIGV evaluation by combining MOS with fine-grained attributions, reducing human review cost and improving reproducibility with an efficient SlowFast design. However, it may be repurposed as a reward model to optimize T2V systems, potentially amplifying deceptive media risks, and subjective labels can introduce bias and reward hacking. We will release Q-Save with clear intended-use guidance and recommend human-in-the-loop deployment, alongside future bias audits, calibration, and robustness studies.

\bibliography{main}
\bibliographystyle{icml2026}

\newpage
\appendix
\onecolumn

\section{Implementation Details}
\label{sec:apdx_impl_details}

We summarize the key hyperparameters used in our experiments:
\begin{itemize}[itemsep=-1pt,left=5pt,topsep=0pt]
    \item \textbf{Video preprocessing:} input FPS (8), slow/fast frame split threshold (95\%), slow/fast resolutions ( $8192 / 2048 \times 32 \times 32$  ).
    \item \textbf{SFT:} optimizer (AdamW), learning rate $\eta_{\text{sft}}$ (5e-6), warmup ratio (0.03), epochs (3), effective batch size(8), CE/MSE loss weights $(\alpha_1,\beta_1)$ (1, 1), and precision (bf16).
    \item \textbf{RL (GRPO):} learning rate $\eta_{\text{rl}}$ (2e-6), batch size (256), number of rollouts $G$ (8), reward weights $\lambda$ (0.5), and training steps (32).
    \item \textbf{Inference:} softmax-based score expectation weights $w_j$ (\( \{1, 0.75, 0.5, 0.25, 0\} \)), and tie threshold for preference prediction (0.05).
\end{itemize}

\section{Evaluation Metrics}
\label{sec:apdx_eval_metrics}
\paragraph{MOS-style scoring (correlation).}
For datasets with per-video MOS labels, we evaluate the agreement between predicted scores $\hat{s}$ and human scores $s$ using Spearman’s rank correlation coefficient (SRCC) and Pearson’s linear correlation coefficient (PLCC). Concretely, $\mathrm{SRCC}=\mathrm{corr}(\mathrm{rank}(\hat{s}), \mathrm{rank}(s))$ measures ranking consistency, while $\mathrm{PLCC}=\mathrm{corr}(\hat{s}, s)$ measures linear agreement.

\paragraph{Preference-style scoring (accuracy).}
For pairwise preference datasets, each test sample contains a pair of videos with a human preference label indicating which video is better (or a tie). Given predicted scores for two videos, we predict the preferred one by comparing their scores; if the score difference is below a small threshold, we predict a tie. Accuracy is computed as the fraction of correctly predicted preference labels. We report both \textit{w/ ties} (including tie cases) and \textit{w/o ties} (excluding tie cases) when the benchmark provides ties.

\paragraph{Instance-level vs. model-level evaluation.}
\textbf{Instance-level} metrics are computed over individual video instances (or video pairs) in the test set. \textbf{Model-level} metrics first aggregate predictions (and ground truths) by the video generator/model identity (e.g., averaging scores over all prompts for each generator), and then compute SRCC/PLCC or accuracy on these aggregated per-model statistics. Model-level evaluation reflects how well an evaluator ranks different generators overall.

\section{Dataset Details}
\label{sec:apdx_dataset_details}

\subsection{Dataset Construction Details}
\label{subsec:apdx_dataset_construction}

\begin{wrapfigure}{r}{0.40\textwidth} 
    \centering
    \includegraphics[width=0.4\textwidth]{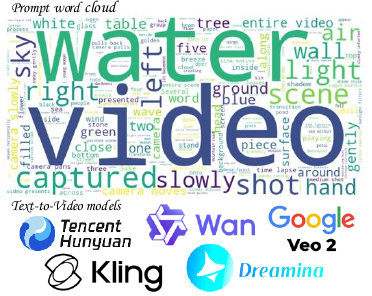}
    \caption{Word cloud of prompts and overview of video generation model sources} 
    \label{prompt_and_model}
\end{wrapfigure}

In this section, we describe the prompt collection process in detail.

\paragraph{Prompts Collection}

Table~\ref{tab:prompt_quantity_range} presents the distribution of prompts across five major entity categories—Human, Animal, Plant, Object, and Virtual Subject—based on the number of times each category appears within a single prompt. The quantity ranges are divided into five bins: 1–3, 4–6, 7–9, 10 or more, and unknown (cases where quantity could not be determined explicitly). Most prompts involving Humans and Objects appear in the 1–3 range, reflecting a common design bias toward fewer instances of these entities. In contrast, categories like Animals and Objects exhibit higher frequencies in the “10 or more” range, likely due to prompts involving groups, crowds, or complex scenes. Notably, the “unknown” bin is most prominent for Plants and Objects, indicating frequent use of implicit or abstract descriptions in those categories. This distribution analysis highlights the variation in prompt complexity and provides insights into how different entity types are quantitatively represented in the dataset.

Table~\ref{tab:classification} presents a detailed breakdown of prompt elements classified into two major categories: \textbf{Scene} and \textbf{Subject}. Each main category is further divided into several subcategories and sub-subcategories to reflect the diversity and granularity of the prompts. For example, the \textbf{Scene} category includes human-centric, natural, and virtual scenes, with further distinctions such as indoor, outdoor, and abstract environments. Similarly, the \textbf{Subject} category encompasses humans, animals, plants, objects, and abstract virtual entities, capturing both physical and conceptual entities. The count column reflects the number of prompt instances associated with each (sub)category, indicating areas with higher representation, such as human subjects (253 instances) and object-related prompts (396 instances). This hierarchical structure helps to systematically analyze prompt diversity and distribution.

\begin{table}[t]
\captionsetup{skip=2pt}
\centering
\caption{Prompt Elements Categorized by Style, Lens Parameters, and Spatial Relationships}
\label{tab:style}
{\scriptsize
\renewcommand{\arraystretch}{1.10}
\setlength{\tabcolsep}{2pt}
\begin{tabular*}{\linewidth}{@{\extracolsep{\fill}} >{\raggedright\arraybackslash}p{0.26\linewidth} >{\raggedright\arraybackslash}p{0.60\linewidth} r}
\toprule
\textbf{Category}                  & \textbf{Subcategory}                            & \textbf{Count} \\ \midrule
\multirow{29}{*}{Style and atmosphere} 
                                   & Atmosphere                     & 109         \\ 
                                   \cline{2-3}
                                   & \quad Sad                  & 16 \\
                                   & \quad Romantic                & 13 \\
                                   & \quad Calm/Tranquil               & 19 \\ 
                                   & \quad Other               & 2 \\
                                   & \quad Fantasy                & 11 \\
                                   & \quad Lively                  & 4 \\
                                   & \quad Warm                  & 4 \\
                                   & \quad Suspenseful                  & 12 \\
                                   & \quad serious                  & 3 \\
                                   & \quad Humorous/Cheerful                & 25 \\
                                   \cline{2-3}
                                   & Style                           & 139           \\
                                   \cline{2-3}
                                   & \quad 3D-cartoon                  & 17         \\
                                   & \quad 80s vapor wave                      & 6        \\
                                   & \quad Master                             & 7    \\
                                   & \quad Anime            & 29    \\  
                                   & \quad Black \& White             & 22   \\   
                                   & \quad Claymation             & 1   \\   
                                   & \quad Moody           & 6  \\   
                                   & \quad Pixel art          & 5   \\   
                                   & \quad Realistic         & 27   \\   
                                   & \quad Sci-fi art         & 9   \\   
                                   & \quad Stickers         & 2   \\   
                                   & \quad Watercolor         & 8   \\   
                                   \cline{2-3}
                                   & Others              &133 \\
                                   \cline{2-3}
                                   & \quad Composition                   & 1         \\
                                   & \quad Light and shadow                    & 67        \\
                                   & \quad intensifiers                           & 31    \\
                                   & \quad color           & 34   \\
                                     \midrule
\multirow{25}{*}{Lens parameters} 
                                   & Scene                          &    113            \\ \cline{2-3}
                                   & \quad Panoramic/Wide angle          & 35          \\
                                   & \quad Close-up                    & 29           \\
                                   & \quad Long shot                        & 22            \\
                                   & \quad Medium shot/Close shot       & 27           \\
                                   \cline{2-3}
                                   & Lens movement             & 168           \\ 
                                   \cline{2-3}
                                   & \quad Tilt                             & 30            \\
                                   & \quad Vertical                              &  21           \\
                                   & \quad Roll                 & 2           \\
                                   & \quad Surround              & 23          \\
                                   & \quad Pan              & 34            \\
                                   & \quad Other              & 8            \\
                                   & \quad Horizontal              & 26           \\
                                   & \quad Push/Pull              & 24            \\
                                   \cline{2-3}
                                   & Shooting techniques                        &    105     \\     
                                   \cline{2-3}
                                   & \quad background blur/depth of field & 2          \\ 
                                   & \quad zoom/scaling            & 22          \\ 
                                   & \quad high-speed photography            & 15           \\ 
                                   & \quad tracking           & 25           \\ 
                                   & \quad slow motion           & 28           \\ 
                                   & \quad time-lapse          & 35           \\ 
                                   \cline{2-3}
                                   & viewing angle                 &   47            \\
                                   \cline{2-3}
                                   & \quad overhead/aerial photography        & 25            \\
                                   & \quad Others                   & 4           \\ 
                                   & \quad looking up             & 18           \\ 
                                     \midrule
\multirow{7}{*}{Spatial relationships}                                    
                                   & Spatial relationships     &    273       \\ 
                                   \cline{2-3}
                                   & \quad Up               & 146         \\
                                   & \quad Down            & 40           \\ 
                                   & \quad Right          & 24          \\ 
                                   & \quad Left            & 23           \\ 
                                   & \quad Back             & 19           \\ 
                                   & \quad Front            & 21           \\ 
\bottomrule
\end{tabular*}
}
\end{table}

\begin{table}[htbp]
\centering
\caption{Categorized Statistics of Prompt Elements Across Scene and Subject Types}
\label{tab:classification}
{\footnotesize
\renewcommand{\arraystretch}{1.05}
\setlength{\tabcolsep}{3pt}
\begin{tabular}{>{\raggedright\arraybackslash}p{0.24\linewidth} >{\raggedright\arraybackslash}p{0.66\linewidth} r}
\toprule
\textbf{Category}                  & \textbf{Subcategory}                            & \textbf{Count} \\ \midrule
\multirow{15}{*}{Scene} 
                                   & Human scenes                       & 136           \\ \cline{2-3}
                                   & \quad Indoor                  & 49 \\
                                   & \quad Outdoor                  & 52 \\
                                   & \quad Famous humanities classics                 & 35 \\ \cline{2-3}
                                   & Virtual Scene                            & 102           \\
                                   \cline{2-3}
                                   & \quad Abstract Scene                    & 18         \\
                                   & \quad Science Fiction Scene                      & 29        \\
                                   & \quad Dream Scene                              & 30    \\
                                   & \quad Virtual and Real Interweaving            & 15    \\  
                                   & \quad Well-known Virtual Scene              & 10   \\   
                                   \cline{2-3}
                                   & Natural Scene              &146 \\
                                   \cline{2-3}
                                   & \quad Air                    & 47         \\
                                   & \quad Land                     & 51        \\
                                   & \quad Water                            & 51    \\
                                   & \quad Famous Natural Attractions            & 26   \\
                                     \midrule
\multirow{33}{*}{Subject} 
                                   & Human                            &    253            \\ \cline{2-3}
                                   & \quad Wear                       & 40          \\
                                   & \quad Age/Era                     & 33           \\
                                   & \quad Ethnicity                         & 28            \\
                                   & \quad Physical Characteristics         & 34           \\
                                   & \quad Number             & 31            \\ 
                                   & \quad Gender                      & 41            \\
                                   & \quad Famous Person                       & 8           \\
                                   & \quad Occupation                      & 38            \\
                                   \cline{2-3}
                                   & Animals/Microorganisms               & 154           \\ 
                                   \cline{2-3}
                                   & \quad Air                              & 49            \\
                                   & \quad Land                              &  51           \\
                                   & \quad Water                  & 52           \\
                                   & \quad Microscopic/internal               & 2            \\
                                   \cline{2-3}
                                   & Plant                        &    63           \\     
                                   \cline{2-3}
                                   & \quad Land             & 32           \\ 
                                   & \quad Water            & 31           \\ 
                                   \cline{2-3}
                                   & Object                                  &   396            \\ \cline{2-3}
                                   & \quad Work/Study Objects                  & 39            \\
                                   & \quad Others                   & 104           \\ 
                                   & \quad Food                 & 62           \\ 
                                   & \quad Cultural/Entertainment Objects          & 46           \\ 
                                   & \quad Travel                 & 37           \\ 
                                   & \quad Clothing                  & 32           \\ 
                                   & \quad Sports/Health                 & 33           \\ 
                                   & \quad Housing                  & 43          \\ 
                                   \cline{2-3}
                                   & Virtual Subject/Abstract Concept     &    227       \\ 
                                   \cline{2-3}
                                   & \quad Abstract symbol                  & 29          \\
                                   & \quad Abstract symbol (text)               & 117           \\ 
                                   & \quad Virtual creature            & 29          \\ 
                                   & \quad Virtual Objects             & 30           \\ 
                                   & \quad Well-known virtual image              & 22           \\ 
                                   \cline{2-3}
                                   & Multi-Object Combination     &    144       \\
\bottomrule
\end{tabular}
}
\end{table}

\begin{table}[t]
\centering
\caption{Distribution of Prompts by Entity Category and Quantity Range. Each cell shows the number of prompts containing a specific category (e.g., Human, Animal) within the indicated quantity range.}
  \label{tab:prompt_quantity_range}
{\small
\renewcommand{\arraystretch}{1.15}
\setlength{\tabcolsep}{6pt}
\begin{tabular}{c|ccccc}
  \hline
  \hline
   \multirow{2}{*}{\textbf{Category}}& \multicolumn{5}{c}{\textbf{Quantity range}} \\
   \cline{2-6}
   & 1-3 & 4-6 & 7-9 & 10 or more & unknown \\
  \hline
  Human & 278 & 25 & 20 & 8 & 7 \\
  Animal & 161 & 49 & 18 & 41 & 15 \\
  Plant & 34 & 16 & 15 & 5 & 76\\
  Object & 434 & 24 & 19 & 65 & 213\\
  Virtual Subject & 82 & 45 & 14& 18 & 13\\
 \hline    
\hline
\end{tabular}
}
\end{table}

Table~\ref{tab:style} provides a comprehensive categorization of prompt-related elements that influence the expressive style and visual dynamics of generated content. The table is divided into three major sections: \textbf{Style and Atmosphere}, \textbf{Lens Parameters}, and \textbf{Spatial Relationships}. The Style and Atmosphere category includes both emotional tones (e.g., sad, humorous, suspenseful) and artistic styles (e.g., anime, watercolor, pixel art), along with other modifiers such as lighting and color intensifiers. The Lens Parameters section captures cinematic techniques used in visual storytelling, such as camera movements (e.g., tilt, pan, push/pull), shooting types (e.g., long shot, close-up), and temporal effects (e.g., slow motion, time-lapse). Finally, the Spatial Relationships category records directional placements (e.g., up, down, front, back), indicating how entities are arranged or oriented in the scene. This taxonomy helps to reveal the diversity and richness of descriptive control available in prompt design for multimodal generation systems.

\paragraph{Subjective Annotation Protocol.}
\label{subsec:apdx_annotation_protocol}
We construct a dataset of 10,000 videos, incorporating rigorous subjective evaluation protocols to ensure annotation quality. Each instance is rated by at least 3 raters in the training set and 12 raters in the test set. For low-quality instances (i.e., those with a mean opinion score (MOS) below 4), we further collect fine-grained \textbf{attribution annotations} to explain the source of quality degradation.

To ensure rater reliability and diversity, we recruit annotators from a wide age range (18–52) and various professional backgrounds. Each rater is limited to labeling a maximum of 30 videos per session, followed by a mandatory 30-minute break to prevent fatigue. For particularly high-quality cases (i.e., perfect-score videos), the annotations are first provided by 12 individuals and subsequently reviewed and adjusted by a group of 5 domain experts.

Because we additionally collect time-consuming attribution annotations for all low-quality videos, following the ITU-recommended standard of 15 raters per instance would be impractical in terms of time and cost. To maintain the dataset’s scale, which is essential for training large multimodal models under scaling laws, we adopt a \textit{Sample and Scrutinize} strategy: a subset of 160 videos is annotated by all raters, with expert-generated ground-truth annotations provided by 10 specialists. Based on the agreement with these ground truths, we identify and exclude the 6 least accurate annotators from the final dataset to ensure labeling quality. The distribution of annotation variance is illustrated in Fig.~9, where the majority of instances exhibit a variance below 0.3, demonstrating strong inter-rater consistency.

\paragraph{Reinforcement Learning Dataset.}
\label{RL_dataset}
We select 200 high-quality prompts from various datasets. These prompts align with our prompt collection principles in terms of difficulty and category distribution, while being distinct from those in our SFT dataset. We use six models that are entirely different from those used for the SFT dataset: Kling2.1, HailuoAI, Veo3, Seedance-1.0-lite, Wanx-2.1-plus, and PixverseV4. The annotation standards and processes for this dataset are identical to those of the SFT dataset.

\subsection{Generative Model Performance}

\begin{figure*}
\centering
\includegraphics[width=1.0\textwidth]{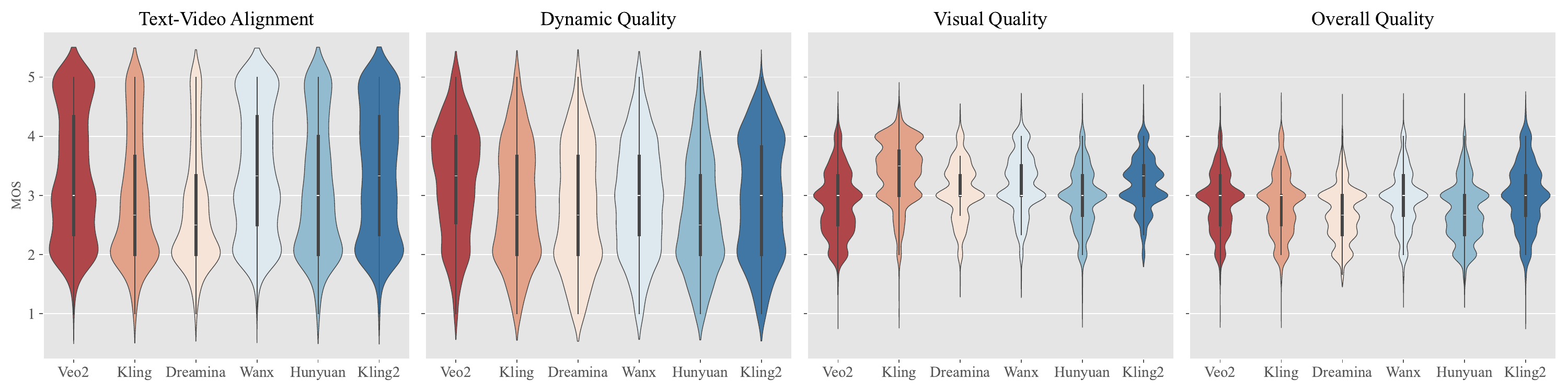}
\caption{MOS distributions for visual quality, dynamic quality, text-video alignment, and overall quality of generated videos in the Q-Save dataset.} 
\label{fig_data}
\end{figure*}

\begin{figure*}
\centering
\includegraphics[width=1.0\textwidth]{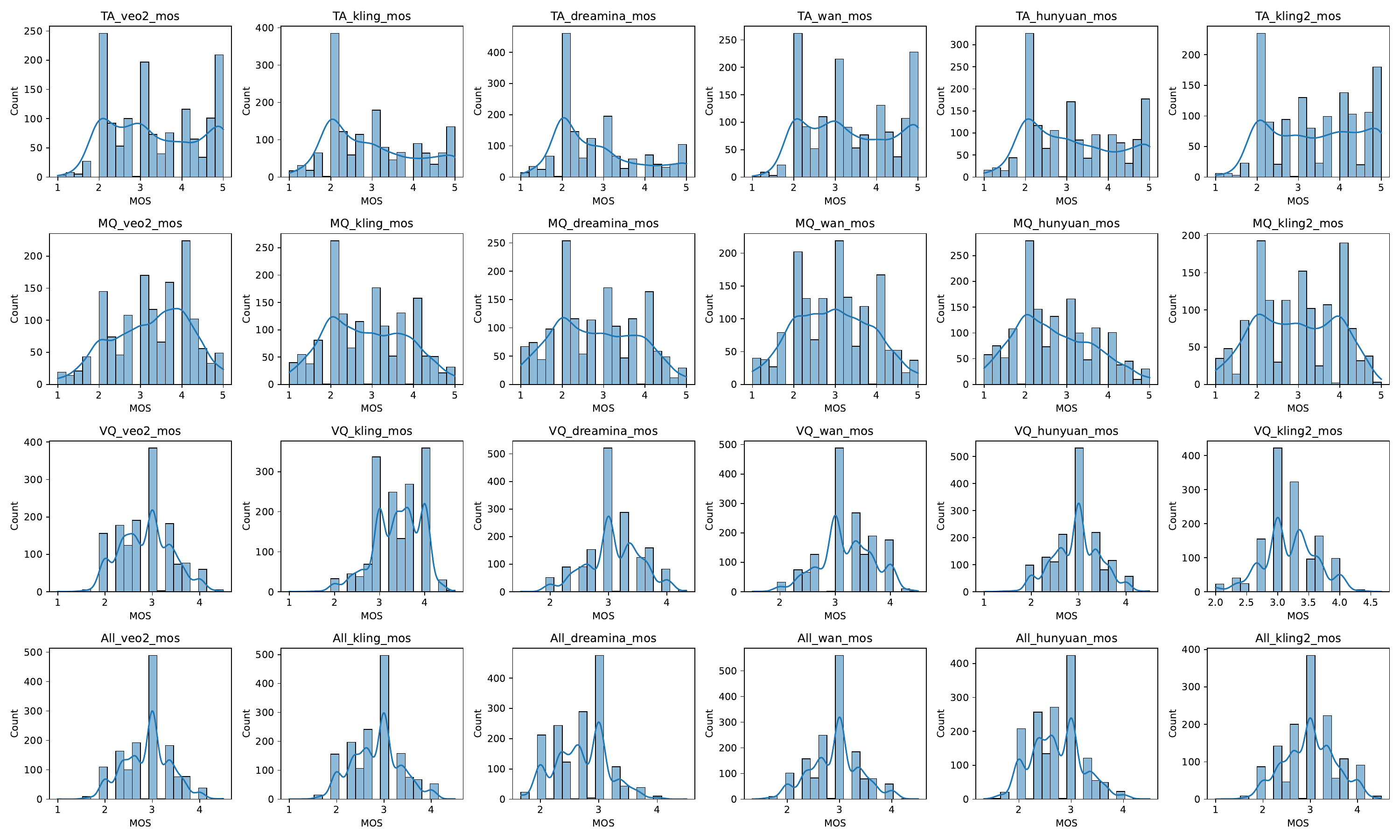}
\caption{Distribution of Mean Opinion Scores (MOS) across six generative models (Veo2, Kling, Kling2.0, Hunyuan, Dreamina, Wan), evaluated along four quality dimensions: Text Alignment (TA), Motion Quality (MQ), Visual Quality (VQ), and Overall Quality (All). Each subplot represents a model-dimension pair, with MOS scores (1–5) on the x-axis and frequency counts (0–500) on the y-axis. Blue bars denote histogram counts, while overlaid curves represent smoothed kernel density estimates.}

\label{mos_distribution}
\end{figure*}

\paragraph{MOS Distribution Analysis.}
Figure~\ref{mos_distribution} visualizes the distribution of Mean Opinion Scores (MOS) for six state-of-the-art generative models—Veo2, Kling, Kling2.0, Hunyuan, Dreamina, and Wan—evaluated across four key quality dimensions: Text Alignment (TA), Motion Quality (MQ), Visual Quality (VQ), and Overall Quality (All). Each subplot in the figure corresponds to a specific model-dimension pair, with histogram bars indicating frequency counts and overlaid curves representing smoothed density estimates. These visualizations provide insights into how different models perform across evaluation axes and semantic domains.

Across models, several distinctive distributional patterns emerge. Veo2 demonstrates strong performance in TA, with scores clustering near 4, while its MQ scores show a bimodal distribution (peaks at MOS 3 and 4.5), suggesting instability in motion generation. Kling2.0 consistently achieves high VQ ratings, with over 80\% of samples rated 4 or higher, but exhibits left-skewed TA scores (median = 3.5), indicating potential issues in text-video alignment. Hunyuan presents a notable anomaly in TA, with a dual-peaked distribution centered around MOS 2.5 and 4—this likely reflects stylistic diversity that polarized raters. In contrast, Wan underperforms across all dimensions, with left-skewed distributions and overall MOS medians between 2.8 and 3.2, suggesting limited robustness in both content generation and coherence.

From a dimension-centric perspective, the TA scores are generally right-skewed (median = 4.1), suggesting that most models are relatively successful in capturing textual alignment, though Hunyuan displays higher variance ($\sigma = 1.2$), pointing to inconsistency. Motion Quality exhibits the greatest inter-model variability: Veo2 shows a bimodal shape, Dreamina’s distribution is nearly uniform (entropy = 1.6), and Kling shows a near-normal shape with moderate variance ($\sigma = 0.7$). Visual Quality is the most predictive of user-perceived quality, with a strong positive correlation with overall scores (Pearson’s $r = 0.89$, $p < 0.001$). Notably, Kling2.0 achieves the most stable overall ratings (95\% confidence interval [4.1, 4.3]), while Dreamina yields the widest spread (IQR = 1.8), suggesting diverse output quality.

Finally, several statistical irregularities are worth noting. Wan’s MQ distribution shows a floor effect, with nearly 30\% of ratings falling below MOS 2, indicating recurrent motion artifacts. Conversely, Veo2’s VQ distribution shows a ceiling effect, with 15\% of samples receiving a perfect score of 5, highlighting superior visual rendering. Additionally, Kling’s TA ratings appear truncated below MOS 2, which may point to filtering biases in sample selection or annotation. These findings collectively reveal both the strengths and limitations of current generative models and motivate further efforts toward dimension-specific evaluation and rater calibration.

\section{More Experimental Details}

\subsection{Reward model baselines}
\label{sec:apdx_reward_model_baselines}

This section provides brief descriptions of the reward-model competitors used in Sec.~\ref{subsec:exp_setup}.

\noindent\textbf{Q-ALIGN} proposes a human-emulating syllabus for training large multimodal models (LMMs) for visual scoring. It converts mean opinion scores (MOS) into five text-defined rating levels (excellent/good/fair/poor/bad) for training, and infers scores via softmax pooling and a weighted average of level probabilities during inference. By unifying image quality assessment (IQA), image aesthetic assessment (IAA), and video quality assessment (VQA) under one LMM framework, it achieves state-of-the-art performance with high data efficiency and strong out-of-distribution generalization.

\noindent\textbf{DOVER} is proposed based on the View Decomposition strategy, with two branches dedicated to the aesthetic and technical perspectives, respectively. It uses specific inputs, regularization strategies, and pretraining to match each perspective’s characteristics, and adopts a subjectively inspired fusion of dual-branch predictions for overall quality evaluation.

\noindent\textbf{VisionReward} introduces a fine-grained, multi-dimensional evaluation framework for both image and video domains. It trains separate reward models tailored to human preferences collected through carefully curated datasets, offering strong baselines for visual content assessment.

\noindent\textbf{VideoScore-v2} focuses on assessing video generation quality. It is trained on the VideoFeedback dataset comprising human-annotated scores over 37.6K videos, each evaluated across multiple aspects including fidelity, consistency, and alignment.

\noindent\textbf{VideoReward} offers a multi-dimensional assessment for video generation tasks. It is trained on a large-scale 182K dataset of human-labeled comparisons collected from outputs of 12 video generation models, providing strong performance on complex video benchmarks.

\noindent\textbf{UnifiedReward} serves as our base architecture. It leverages multi-task learning across diverse image and video generation and understanding datasets. By unifying multimodal reward tasks into a single framework, UnifiedReward demonstrates mutual enhancement effects and establishes a solid baseline for holistic visual reward modeling.

\noindent\textbf{AIGVE-MACS} is a unified evaluation framework for AI-generated videos. It is fine-tuned on the large-scale human-annotated AIGVE-BENCH 2 dataset, integrating token-wise weighted loss and a dynamic frame sampling strategy to fully leverage the generative capabilities of vision-language models (VLMs). By jointly producing aspect-wise numerical scores and multi-aspect natural language comments, it achieves strong alignment with human judgments and sets a new paradigm for interpretable and comprehensive AI-generated video evaluation.

\noindent\textbf{VIDEOPHY-2} is an action-centric benchmark for evaluating physical commonsense in video generation. It curates 197 diverse real-world actions, generates 3940 detailed prompts (including dense upsampled versions), and constructs a hard subset for rigorous testing. Through human evaluation on semantic adherence, physical commonsense, and fine-grained physical rule compliance, it exposes model shortcomings, especially in conservation laws. Complemented by the VIDEOPHY-2-AUTOEVAL automatic evaluator (fine-tuned on 50K human annotations), it provides a scalable, reliable framework for guiding physically grounded video generation research.

\subsection{Evaluation benchmarks}
\label{subsec:apdx_eval_benchmarks}

This section describes the out-of-domain benchmarks used in the cross-dataset validation in Sec.~\ref{subsec:discussion_findings}.

Since different benchmarks define varying dimensions and scoring scales, we align them with the three evaluation dimensions of \textbf{Q-Save} (visual quality, text alignment, and physical consistency) and, where necessary, rescale their ground-truth scores. 

\paragraph{VideoGenReward Bench.}
This is a pairwise preference benchmark containing 4,691 videos, forming 25,234 video pairs. It evaluates three dimensions---visual quality (VQ), text alignment (TA), and motion quality (MQ)---and also provides an \textit{Overall} preference label indicating which video is better overall. For the \textit{Overall} preference, we use the mean of all available dimension scores from Q-Save or the baseline method (if the baseline only has one quality score output, then that score is used directly).

\paragraph{T2VQA-DB.}
T2VQA-DB is a human-annotated video quality dataset with 10,000 videos, each labeled with a single quality score in the range [1,100]. We sample 2,000 videos and construct 1,822 pairs by comparing human-annotated scores. Since the dataset provides only one dimension (the final score), we predict preference by averaging all dimension scores from Q-Save or the baseline method (if the baseline only has one quality score output, then that score is used directly).

\paragraph{GenAI-Bench.}
This is a multimodal generation benchmark designed to assess how well models align with human preferences across image and video generation tasks. We adopt its video generation subsets for evaluating generative reward performance.

\paragraph{VideoPhy2-test.}
This benchmark contains 3,396 videos with two dimensions, SA: \textit{semantic adherence} and PC:  \textit{physical consistency}. These map perfectly to Q-Save’s second and third dimensions. For baselines lacking one of the dimensions (e.g., VideoReward, which provides VQ, TA, MQ but no physical consistency), we skip the missing dimension. The scoring scale is \{1,2,3,4,5\}, so no rescaling is required.

\paragraph{VIDEOSCORE-BENCH-V2. }
This is a point-score benchmark constructed from 500 held-out videos in the VIDEOFEEDBACK2 dataset. It evaluates videos across three core dimensions: visual quality, text alignment, and physical/common-sense consistency. The benchmark uses integer scores in the range [1,5] and adopts multiple evaluation metrics, including Accuracy (exact match between model and human scores), Relaxed Accuracy (allowing a maximum 1-point difference from human scores), and PLCC (Pearson Linear Correlation Coefficient) to measure the consistency between model predictions and human annotations.

\subsection{Full evaluation results}
\label{sec:apdx_full_eval_results}

\subsubsection{Full results on VideoGen-RewardBench}

VideoGen-Reward-Bench is a video preference benchmark over three dimensions: visual quality, text alignment, and motion quality. The task is to compare a pair of videos and judge which one is better along these axes. Among them, the first two dimensions are broadly aligned with ours, while the benchmark also provides an additional measure of overall preference.

For the preference benchmarks, we report results under two settings. The w/ Ties version includes all test entries, where in some cases the two compared videos (including the ground-truth reference) are judged as equally preferred. The w/o Ties version is a subset obtained by removing those entries with equal preference labels. The full evaluation results of preference prediction accuracy are shown in Table~\ref{tab:apdx_full_res_vgrbench}.

\begin{table*}[htbp]
  \captionsetup{skip=3pt}
  \caption{Full evaluation results on \textbf{VideoGen-Reward-Bench}. \textbf{Bold} denotes the best model and the \underline{underlined} denotes the second best.}
  \label{tab:apdx_full_res_vgrbench}
  \centering
  \small
  \renewcommand{\arraystretch}{1.3}
  \setlength{\tabcolsep}{4pt}
  \begin{tabular}{
    >{\centering\arraybackslash}p{90pt}|
    >{\centering\arraybackslash}p{45pt}|
    >{\centering\arraybackslash}p{45pt}|
    >{\centering\arraybackslash}p{45pt}|
    >{\centering\arraybackslash}p{45pt}|
    >{\centering\arraybackslash}p{45pt}|
    >{\centering\arraybackslash}p{45pt}
  }
    \hline
    \hline
    \multirow{2}{*}{\textbf{Method}} & \multicolumn{2}{c|}{\textbf{Visual Quality}} & \multicolumn{2}{c|}{\textbf{Text Alignment}} & \multicolumn{2}{c}{\textbf{Overall}} \\
    \cline{2-7}
    & w/ Ties & w/o Ties & w/ Ties & w/o Ties & w/ Ties & w/o Ties \\
    \hline
    VisionReward & 35.89 & 59.03 & \underline{44.86} & \underline{61.15} & \underline{54.31} & \underline{67.58} \\
    Q-Align & 32.01 & 52.98 & 35.77 & 51.06 & 42.05 & 52.52 \\
    AIGVE-MACS & 38.05 & 30.80 & 30.76 & 11.66 & 37.09 & 37.08 \\
    VideoPhy2 & - & - & 37.04 & 22.14 & 30.75 & 26.41 \\
    Dover & \underline{39.34} & \textbf{68.87} & 38.01 & 55.65 & 54.27 & \textbf{68.58} \\
    VideoScore-v2 & 37.44 & 63.08 & 42.87 & 60.61 & 51.53 & 63.72 \\
    Q-Save & \textbf{40.34} & \underline{67.49} & \textbf{50.52} & \textbf{63.99} & \textbf{56.63} & 65.13 \\
    \hline
    \hline
  \end{tabular}
\end{table*}

\subsubsection{Full results on Video-Phy2-test}

Video-Phy2-Test is a human-annotated test set with two dimensions: semantic adherence and physical consistency (abbreviated as semantic and physical in our tables). These two dimensions correspond directly to the latter two evaluation dimensions in our framework.

The full evaluation results on Video-Phy2-Test are shown in Table~\ref{tab:apdx_full_res_videophy2}, with PLCC between model outputs and ground truths adopted as metrics.

\begin{table*}[htbp]
  \captionsetup{skip=3pt}
  \caption{Full evaluation results on \textbf{Video-Phy2-test}. \textbf{Bold} denotes the best model and the \underline{underlined} denotes the second best.}
  \label{tab:apdx_full_res_videophy2}
  \centering
  \small
  \renewcommand{\arraystretch}{1.3}
  \setlength{\tabcolsep}{4pt}
  \begin{tabular}{
    >{\centering\arraybackslash}p{90pt}|
    >{\centering\arraybackslash}p{45pt}|
    >{\centering\arraybackslash}p{45pt}|
    >{\centering\arraybackslash}p{45pt}
  }
    \hline
    \hline
     \multirow{2}{*}{\textbf{Method}} & \multicolumn{3}{c}{\textbf{PLCC}} \\
    \cline{2-4}
         & \textbf{Semantic} & \textbf{Physical} & \textbf{Avg} \\
    \hline
    VisionReward & 28.11 & 13.67 & 20.89  \\
    Q-Align & 5.52 & 2.70 & 4.11  \\
    AIGVE-MACS & 8.09 & 11.90 & 10.00  \\
    VideoPhy2 & 38.64 & 29.84 & 34.24  \\
    Dover & 3.85 & 1.15 & 2.50  \\
    VideoScore-v2 & 41.08 & 17.57 & 29.33  \\
    Q-Save & \textbf{48.34} & \textbf{34.02} & \textbf{41.18}  \\
    \hline
    \hline
  \end{tabular}
\end{table*}

\subsubsection{Human Evaluation Protocol for RL Fine-tuning.}
\label{subsec:apdx_rl_human_eval}
For the human evaluation reported in Table~\ref{tab:rl_video_gen}, a total of 15 domain experts participated, rating 170 prompts. Each prompt was independently evaluated by three experts. We adopt a five-level ordinal scale ({Bad, Poor, Fair, Good, Excellent}), and report the averaged score across raters and prompts. For ease of presentation, we convert the five levels to a percentage scale by a linear mapping (Bad/Poor/Fair/Good/Excellent $\rightarrow$ 0/25/50/75/100). 

\section{Visualization Results}

\begin{figure*}
\centering
\includegraphics[width=1.0\textwidth]{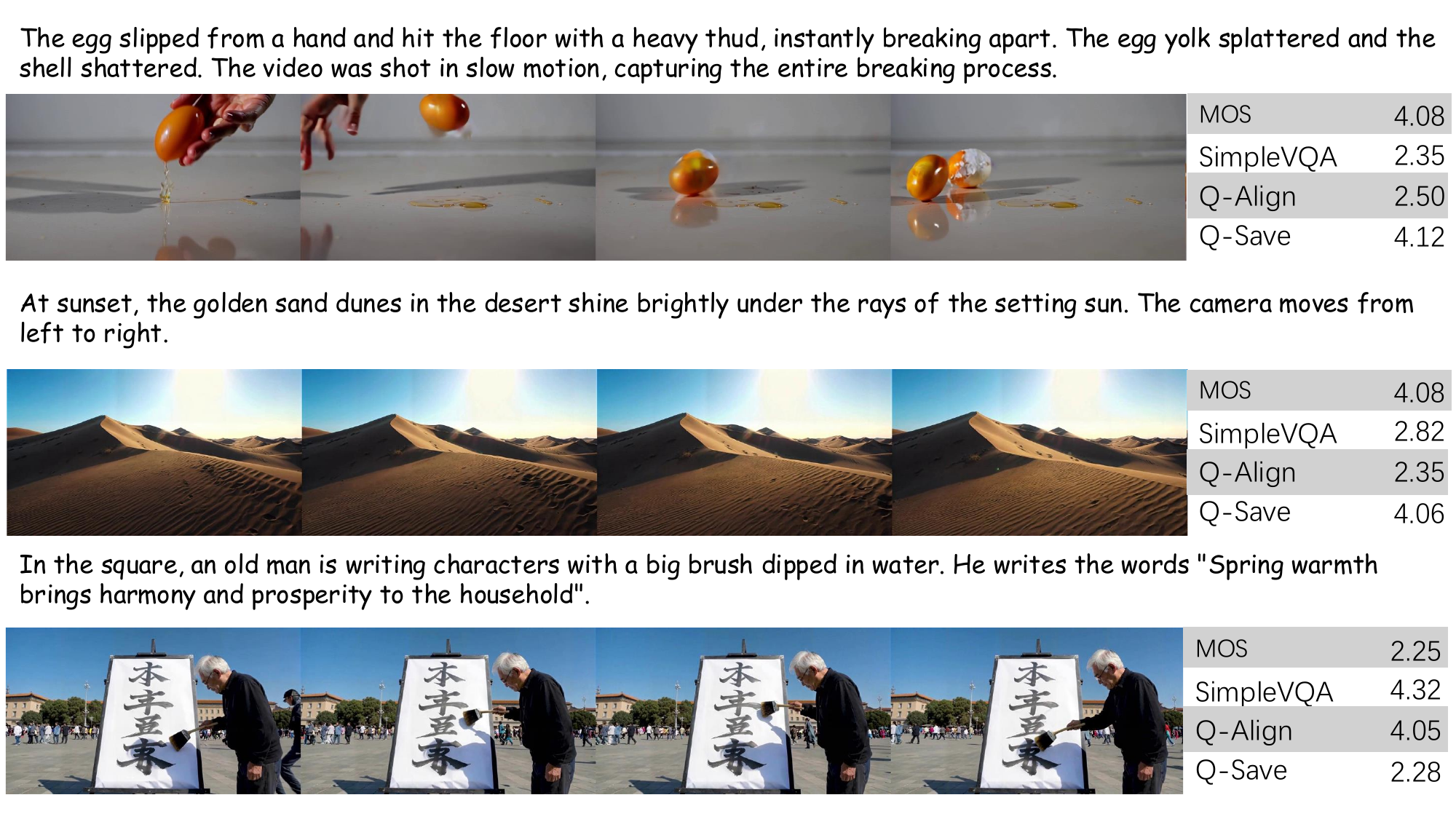}
\caption{Qualitative comparison of text-video alignment.}
\label{visual_result_TA}
\end{figure*}

\begin{figure*}
\centering
\includegraphics[width=1.0\textwidth]{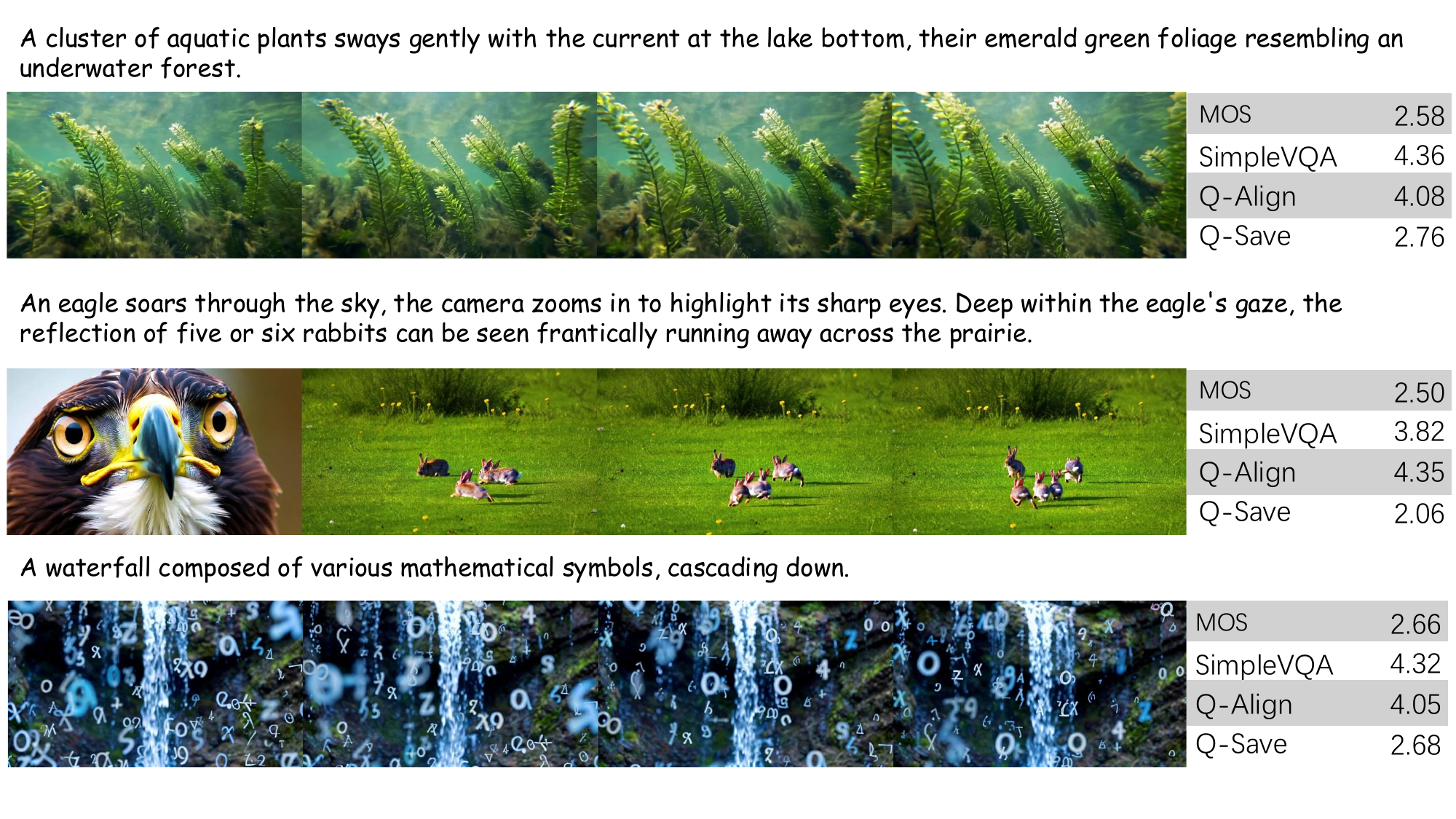}
\caption{Qualitative comparison of visual quality.}
\label{visual_result_VQ}
\end{figure*}

\begin{figure*}
\centering
\includegraphics[width=1.0\textwidth]{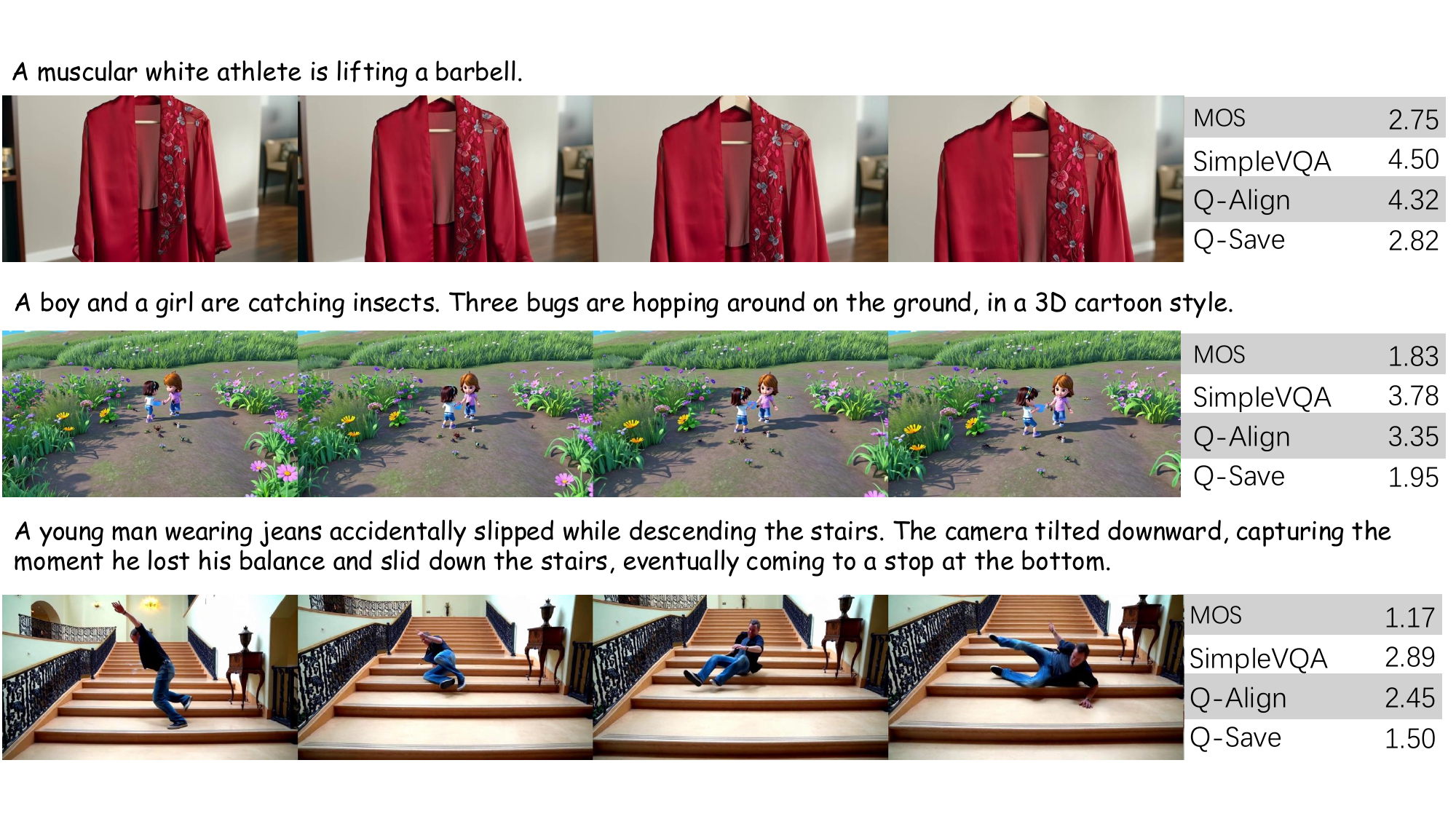}
\caption{Qualitative comparison of dynamic quality.}
\label{visual_result_DQ}
\end{figure*}

\begin{figure*}
\centering
\includegraphics[width=1.0\textwidth]{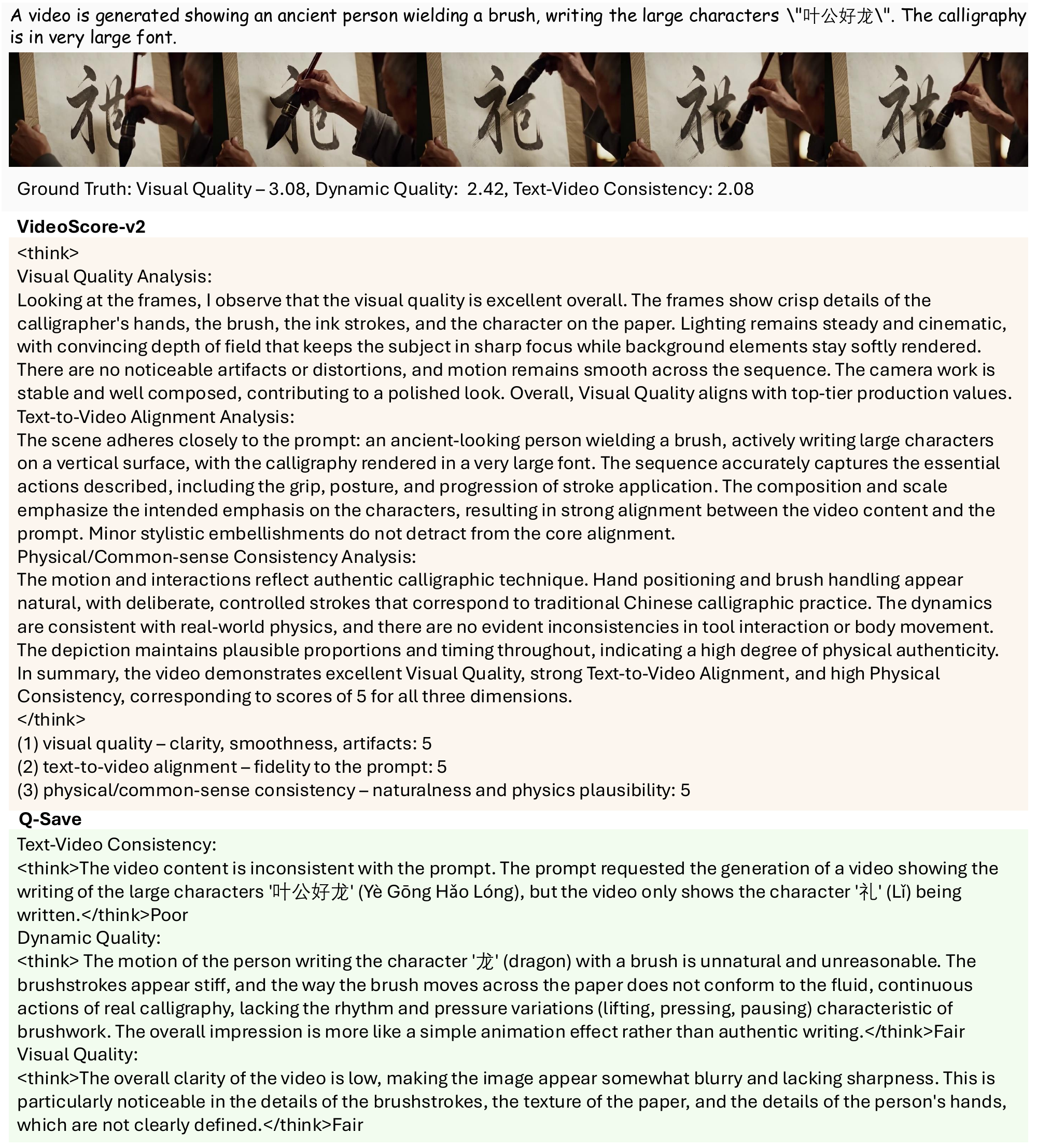}
\caption{Qualitative comparison of CoT outputs.}
\label{case_cot}
\end{figure*}

Using MOS as the ground truth, we summarize evaluator performance across three dimensions (three examples each) in Figures~\ref{visual_result_TA}, \ref{visual_result_VQ}, \ref{visual_result_DQ}, and \ref{case_cot}. Overall, Q-Save performs consistently across dimensions, supporting its value as a unified text-to-video evaluation framework.

\section{Data Statement}

Given the scale of the dataset and the complexity of the model, we are organizing and refining the release to ensure quality and usability. We will release the \textbf{Q-Save} dataset through a structured open-source process that supports community development. \textbf{Furthermore, we confirm that the dataset has passed ethical review, affirming our commitment to responsible AI practices.} Alongside the dataset, we will also release the \textbf{Q-Save} code and model checkpoints, and provide updates to keep pace with rapid advances in generative AI.

\section{Limitations and Future Work}
\label{sec:limitations_and_future_work}

We have not yet provided a thorough analysis of failure cases, such as when the model assigns incorrect scores or produces irrelevant explanations, or whether the attributions exhibit systematic biases. In future work, we plan to (i) curate and release a structured failure set covering challenging motion/physics cases, style shifts, and prompt ambiguity; (ii) analyze bias patterns (e.g., over-penalizing specific styles or over-attributing errors to a single cue); and (iii) improve robustness via targeted data augmentation and calibration techniques.

\end{document}